\documentclass[10pt,twocolumn,letterpaper]{article}

\usepackage{iccv}
\usepackage{times}
\usepackage{epsfig}
\usepackage{graphicx}
\usepackage{amsmath}
\usepackage{amssymb}

\usepackage{multicol}
\usepackage{multirow}
\graphicspath{{./Images/}}
\graphicspath{{./Images_Supp/}}
\usepackage{dsfont}
\usepackage{mathtools}
\usepackage{subcaption}
\usepackage{siunitx}
\usepackage{caption}
\newcommand\tab[1][1cm]{\hspace*{#1}}
\DeclareMathOperator*{\argmax}{argmax} 

\usepackage[pagebackref=true,breaklinks=true,letterpaper=true,colorlinks,bookmarks=false]{hyperref}

\iccvfinalcopy 

\def\iccvPaperID{6663} 

\ificcvfinal\fi

\begin{document}

\title{Adaptive Pseudo-Label Refinement by Negative Ensemble Learning for Source-Free Unsupervised Domain Adaptation}


\author{
Waqar Ahmed$^{1,2}$, Pietro Morerio$^{1,3}$ and Vittorio Murino$^{1,3,4}$\\
$^{1}$Pattern Analysis \& Computer Vision (PAVIS), Istituto Italiano di Tecnologia, Genova, Italy\\
$^{2}$Dipartimento di Ingegneria Navale, Elettrica, Elettronica e delle Telecomunicazioni,\\
University of Genova, Italy\\
$^{3}$Ireland Research Center, Huawei Technologies Co. Ltd., Dublin, Ireland\\
$^{4}$Dipartimento di Informatica, University of Verona, Italy\\
{\tt\small \{waqar.ahmed, pietro.morerio, vittorio.murino\}@iit.it}
}

\maketitle
\ificcvfinal\thispagestyle{empty}\fi

\begin{abstract}
The majority of existing Unsupervised Domain Adaptation (UDA) methods presumes source and target domain data to be simultaneously available during training. Such an assumption may not hold in practice, as source data is often inaccessible (e.g., due to privacy reasons). On the contrary, a  pre-trained source model is always considered to be available, even though performing poorly on target due to the well-known domain shift problem. This translates into a significant amount of misclassifications, which can be interpreted as structured noise affecting the inferred target pseudo-labels. In this work, we cast UDA as a pseudo-label refinery problem in the challenging source-free scenario. We propose a unified method to tackle adaptive noise filtering and pseudo-label refinement. A novel Negative Ensemble Learning technique is devised to specifically address noise in pseudo-labels, by enhancing diversity in ensemble members with different stochastic (i) input augmentation and (ii) feedback. In particular, the latter is achieved by leveraging the novel concept of Disjoint Residual Labels, which allow diverse information to be fed to the different members. A single target model is eventually trained with the refined pseudo-labels, which leads to a robust performance on the target domain. Extensive experiments show that the proposed method, named Adaptive Pseudo-Label Refinement, achieves state-of-the-art performance on major UDA benchmarks, such as Digit5, PACS, Visda-C, and DomainNet, without using source data at all.
\end{abstract}

\vspace{-0.8em}
\section{Introduction}         
\vspace{-0.1em}
Deep Convolutional Neural Networks (CNNs) have shown remarkable achievements in a variety of tasks \cite{dargan2019survey}.
However, to perform well, training and testing data are assumed to be drawn from the same distribution. This is unrealistic when the system needs to be deployed in real-world scenarios. Consequently, a model trained on some source domain often fails to generalize well on a related but different target domain, due to the well-known \emph{domain shift} \cite{torralba2011unbiased} problem. 
\begin{figure}[!t]
\setlength{\belowcaptionskip}{-5pt}
\centering
\includegraphics[width=\linewidth]{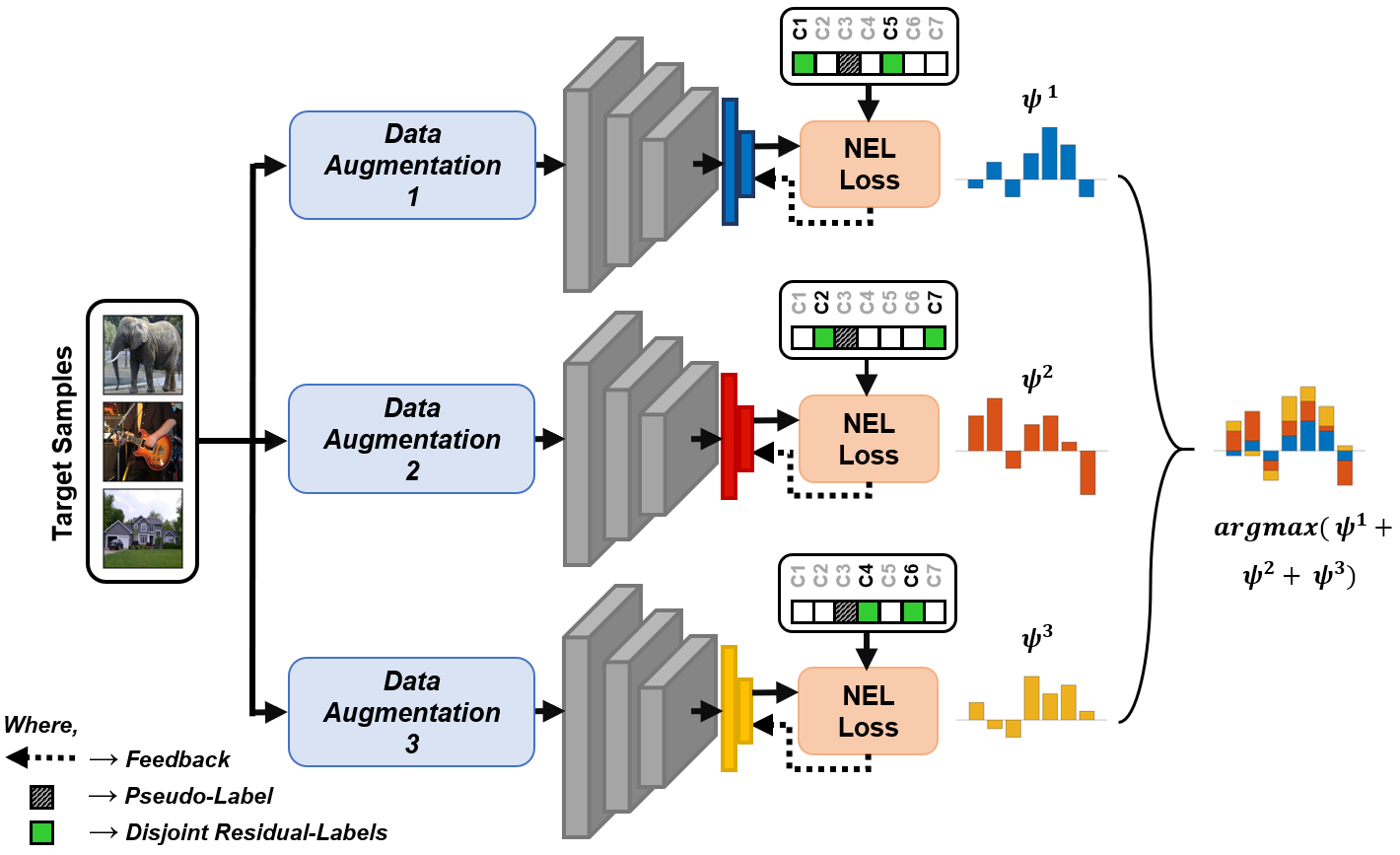}
\caption{\small{Illustration of the proposed unified technique for adaptive noise filtering and progressive pseudo-label refinement (AdaPLR).
After inferring target pseudo-labels using the source model, each ensemble member is trained using a batch of target samples with different stochastic (i) input augmentation and (ii) feedback, by leveraging  \textit{Disjoint residual labels} and \textit{Negative Ensemble Learning} (NEL) loss. 
}}
\vspace{-0.5em}
\label{fig:pipeline}
\end{figure}
Since annotating data from every possible domain is expensive and sometimes even impossible, Unsupervised Domain Adaptation (UDA) methods seek to address such a problem by minimizing discrepancy across the domains or trying to learn domain-invariant feature embeddings, without accessing target label information.
Most UDA works focus on the single-source/single-target scenario, regardless of the fact that data may belong to multiple source or target distributions, \eg, images taken in different environments or obtained from the web (such as sketches, photos, etc.).
Our framework naturally allows to cope with any number of source and target domains, unlike previous literature.

Several research efforts have been devoted to developing UDA methods by either enforcing class-level feature distribution alignment \cite{chen2019progressive, cicek2019unsupervised}, 
matching moments \cite{peng2019moment, chen2020homm},
applying domain-specific batch normalization \cite{chang2019domain}, or 
adopting domain adversarial learning \cite{tang2020discriminative}.
However, all of these methods require joint access to both (labeled) source and (unlabeled)  target data during training, making them unsuitable for scenarios where source data is inaccessible during the adaptation stage, or when source and target data are not available at the same time. Further, such solutions are also not viable when target data is provided incrementally at different times or if the source/target datasets are very large.
%

In this paper, to get rid of such restrictive assumptions, we propose to address UDA from a novel perspective, \ie, by casting it as a pseudo-label refinement problem in a source data-free scenario.
In particular, our proposed framework assumes a model pre-trained on the source domain to infer \emph{pseudo-labels} of unlabeled target samples. This results in a significant amount of incorrect pseudo-labels, which we interpret as \emph{shift-noise} \cite{morerio2020generative} affecting the ground-truth target data labels.
To clean such labels, we propose Negative Ensemble Learning, a novel pipeline where each member of an ensemble is trained using a batch of target samples with different stochastic input augmentation together with the novel concept of disjoint-feedback using \textit{Disjoint Residual Labels}. 

The intuition behind using an ensemble is that in the case of wrong pseudo-labels at least $N_e-1$ members out of $N_e$ can receive the correct information. This is made possible by the fact that in Negative Learning only the complement (or residual) of the original label is used as feedback for the networks. Further, stochastically sampled disjoint subsets of residual labels force the ensemble members to learn different concepts, which is known to be beneficial in ensemble learning to reach a strong consensus.
A strong consensus also achieves a higher probability on clean pseudo-labels, allowing us to introduce a novel \textit{fully adaptive} noise filtering technique which allows refining labels of the samples with low confidence (\ie, noisy pseudo-labels) via reassignment. 
Finally, a standard supervised learning procedure is used to train a \textit{single} model on the target domain data with the refined pseudo-labels featuring high confidence only.

The proposed pipeline obtains significant noise reduction from the inferred pseudo-labels. With extensive experiments on benchmarks with diverse complexity and issues, we show that training on the target domain with refined pseudo-labels outperforms state-of-the-art UDA methods by a considerable margin. 
To summarise, the contributions of our work can be stated as follows.
\begin{itemize} 
\setlength\itemsep{0pt}
  \item We are the first to 
  cast the source-free UDA problem in the framework of label noise reduction. Further, we are proposing a method that can cope with single-source, multi-source and multi-target UDA indifferently, unlike most of the former works. 
  \item We propose a new, fully-adaptive method that dynamically filters out label noise and assign cleaner pseudo-labels to noisy target samples. To do so, we introduce  Negative Ensemble Learning, a new strategy that enhances diversity among members, which improve noise resilience and leads to a stronger consensus. 
 
\item We validate our proposed method through detailed ablation analyses and extensive experiments on four well-known public benchmarks, demonstrating its superiority over state-of-the-art UDA methods
(with a large margin of up to 21.8$\%$ for PACS benchmark).
\end{itemize}

The remainder of the paper is organized as follows. In Section~\ref{RW}, we discuss related works. Section~\ref{Method} describes the proposed method to tackle source-free UDA. Section~\ref{EXP} illustrates the experimental setup and reports the obtained results. Finally, the conclusions are drawn in Section~\ref{Conclusion}.

\section{Related Work}\label{RW}
\textbf{Unsupervised Domain Adaptation.}
Most of the existing UDA methods focus on cross-domain feature \textit{alignment}, either by employing sampling-based implicit alignment \cite{jiang2020implicit}, locality preserving projection based subspace alignment \cite{minnehan2019deep}, discriminative class-conditional alignment \cite{cicek2019unsupervised}, features and prototype alignment using reliable samples \cite{chen2019progressive}, or a customized CNN with domain alignment layers and feature whitening \cite{roy2019unsupervised}. 
Some other works propose feature distribution \emph{matching} by 
approximating joint distribution \cite{ijcai2019-534}, graph matching \cite{das2018graph}, 
moment matching \cite{peng2019moment}, and higher-order moment matching of non-Gaussian distributions\cite{chen2020homm}.
However, such methods assume co-existence of source and target data during training, making them unsuitable for more realistic scenarios where source data is inaccessible, e.g., due to data-privacy issues.

\textbf{Source-free UDA.}
Some recent works focus on the source-free UDA. 
For instance, \cite{chidlovskii2016domain} proposes a feature corruption and marginalization technique using few labeled source samples as representative examples, and \cite{nelakurthi2018source} adapts the outputs from an off-the-shelf model to minimize label deficiency and distribution shift using some labeled target samples.
An instance-level weighting method using negative classes is proposed in \cite{Kundu_2020_CVPR} which is highly dependent on procurement stage which requires the source data.
Another approach leverages a pre-trained source model to update the target model progressively
by generating target-style samples through conditional generative adversarial networks \cite{morerio2020generative}, also combined with clustering-based regularization \cite{li2020model}. Similarly, to improve UDA performance in person re-identification task, \cite{ge2020mutual} proposes a pseudo-label cleaning process with on-line refined soft pseudo-labels.

Our proposed approach lies in this category and partially takes inspiration from the approaches developed for source-free UDA in \cite{li2020model,morerio2020generative}. In both methods, a pre-trained source model is used to infer pseudo-labels of target data, and then exploit a target-style sample generator for adaptation. We instead tackle source-free UDA from a novel perspective, namely by progressively refining  pseudo-labels exploiting the consensus of an ensemble network, without generating any (target-style) data. 

\textbf{Ensemble Learning Methods.}
Such methods exploit features extracted from multiple models (typically CNNs) through a diversity of projections on data, and bring forward the mutual consensus to achieve better performances than those obtained by any individual model  \cite{zhou2012ensemble}.
A comprehensive review about ensemble methods is well presented in \cite{dong2020survey}. 
The concept of diversity contribution ability for classifier selection and weight adjustment is proposed in \cite{yin2014novel}. Also, multiple choice learning is employed in \cite{garcia2019dmcl} to improve the accuracy of an ensemble of models. 

We draw inspiration from the general idea proposed in these works, which agree in stressing that \textit{diversity} among members is beneficial for ensemble robustness. We differentiate from them by introducing a novel way of inducing diversity in the members, \ie we back-propagate different feedback to each member by leveraging the novel concept of Disjoint Residual Labels. This allows each member to learn something diverse and possibly complementary, leading to a stronger consensus and noise resilience. 

\textbf{Learning with Noisy Labels.}
Deep CNNs are capable of memorizing the entire data even when labels are noisy \cite{kim2019nlnl}.
To overcome such overfitting over noisy samples, existing methods try to select a subset of possibly clean labels for training, \eg, using two networks under a co-teaching framework \cite{han2018co}, meta-learning based exemplar weight estimation \cite{zhang2020distilling}, one-out filtering approach based on the local and global consistency algorithm \cite{de2020identifying}, or Negative Learning (NL) as an indirect learning method \cite{kim2019nlnl}.

However, these works only consider random noise from selective or uniform distribution which has a completely different structure from the domain-shift noise associated with the inferred pseudo-labels (see Section~\ref{BG}). Our work is inspired by \cite{kim2019nlnl} which presents a Negative Learning (NL) approach to clean 
samples with noisy labels. Nevertheless, \cite{kim2019nlnl} fails when the noise is not uniform, and the performance is actually affected by threshold sensitivity, which limit the generalization capability of the method across benchmarks.
In contrast to a fixed threshold, our method features a fully \textit{adaptive} procedure to progressively filter out the actual (structured) noise affecting target pseudo-labels under domain shift (\textit{shift-noise} \cite{morerio2020generative}): as the ensemble gets more and more confident in the clean examples during training, label reassignment becomes more rigorous, which prevents the network to overfit noisy samples.

\section{The Method}\label{Method}
In this section, we first review the background notions behind the employed techniques -- Negative Learning and Ensemble Learning -- highlighting their limitations along with the proposed advancements. Then, we illustrate the problem setup and detail the proposed AdaPLR method.

\subsection{Background and Notation}\label{BG}
\vspace{-0.4em}
\textit{\textbf{Negative Learning (NL).}} It was proposed by \cite{kim2019nlnl} which refers to an indirect learning method that leverages inverse supervision. In short, the feedback given to a classifier does not follow the usual "this is class $c$" logic, but rather a "this is \textit{not} class $\bar{c}$ " scheme. 
The intuition is that, by sampling $\bar{c}$ from the complement set of $c$, there are less chances of providing wrong feedback to the network in case the label $c$ is incorrect.

In this paper, we schematize classical NL in the context of UDA for a $C$-class classification task as follows:
\begin{enumerate}
\setlength\itemsep{0pt}
    \item We use a source model to infer pseudo-labels for the whole target set $\mathcal{D}_t$. Such set of labels will be noisy.
    \item For a given sample $\boldsymbol{x}$ with inferred pseudo-label $\tilde{y}$, we train the network with a complementary label $\bar{y}$ (randomly selected from $\{1,...,C\} \backslash \{\tilde{y}\}$) by \textit{minimizing} the probability of $\boldsymbol{x}$ belonging to $\bar{y}$.
\end{enumerate}

Note that the the standard procedure would require \textit{maximizing} the probability of $\boldsymbol{x}$ belonging to $\tilde{y}$. In case of wrongly inferred pseudo-label $\tilde{y} \neq y_t$ (where $y_t$ is the inaccessible actual target label) the network would undeniably be provided the wrong information. Instead, our schematized NL approach would reduce such probability to $\frac{1}{(C-1)}$. 

Let our CNN architecture be composed of a feature extractor $\nu_{\boldsymbol{\phi}}(.)$, a classifier $\psi_{\boldsymbol{\theta}}(.)$, and a \textit{softmax} $\sigma(.)$, being $\boldsymbol{\phi}$ and $\boldsymbol{\theta}$ the related network parameters. 
The function $f: \mathcal{X} \to \mathbb{R}^{C}$, defined as $f(\boldsymbol{x})=\sigma(\psi_{\boldsymbol{\phi}}(\nu_{\boldsymbol{\theta}}(\boldsymbol{x})))$\footnote{Networks' parameters $\theta$ and $\phi$ will be omitted for brevity from now on.}, maps the input  $\boldsymbol{x} \in \mathcal{X}$ to the $C$-dimensional vector of probabilities $\boldsymbol{p} \in \mathbb{R}^{C}$.
Within a standard training procedure, namely Positive Learning (PL), the cross entropy loss function can be defined as:
\vspace{-0.4em}
\begin{equation}\label{eq:pl}
\mathcal{L}_{PL}(\mathcal{D}_t)=-\mathbb{E}_{x_t \sim \mathcal{D}_t}\sum_{c=1}^{C} \mathds{1}_{[c=\tilde{y}]}\log(\boldsymbol{p})
\end{equation}

\noindent
where $\mathds{1}$ is an indicator function, $\mathcal{D}_t$ represents the unlabeled target domain and $\boldsymbol{p} = f(\boldsymbol{x}_t )$. Clearly, Eq.~(\ref{eq:pl}) pushes the probability $p$ for the given pseudo-label $\tilde{y}$ towards $\boldsymbol{p}_{\tilde{y}} = 1$. On the contrary, NL aims at  encouraging the probabilities of complementary labels $\bar{y}$ to move  
away from $1$, actually pushing them towards $\boldsymbol{p}_{\bar{y}} = 0$. The NL loss function would be so defined as:
\begin{equation}\label{eq:nl}
\mathcal{L}_{NL}(\mathcal{D}_t)=-\mathbb{E}_{x_t \sim \mathcal{D}_t}\sum_{c=1}^{C} \mathds{1}_{[c=\bar{y}]}\log(1-\boldsymbol{p})
\end{equation}

\noindent
where $\bar{y}$ is selected randomly from $\{1,...,C\} \backslash \{\tilde{y}\}$ at every training iteration. While Eq.~(\ref{eq:nl}) optimizes the output probability of the complementary label to be close to zero, the probability values of other classes are increased. In such a contest, the samples carrying clean $\tilde{y}$ get higher confidence, whereas noisy ones struggle with scarcely confident scores, meeting the purpose of NL. 
\begin{figure}[!t]
\setlength{\belowcaptionskip}{-5pt}
\captionsetup{font=small}
\centering
\begin{subfigure}{0.49\linewidth}
\centering
\captionsetup{font=footnotesize}
\includegraphics[width=\linewidth]{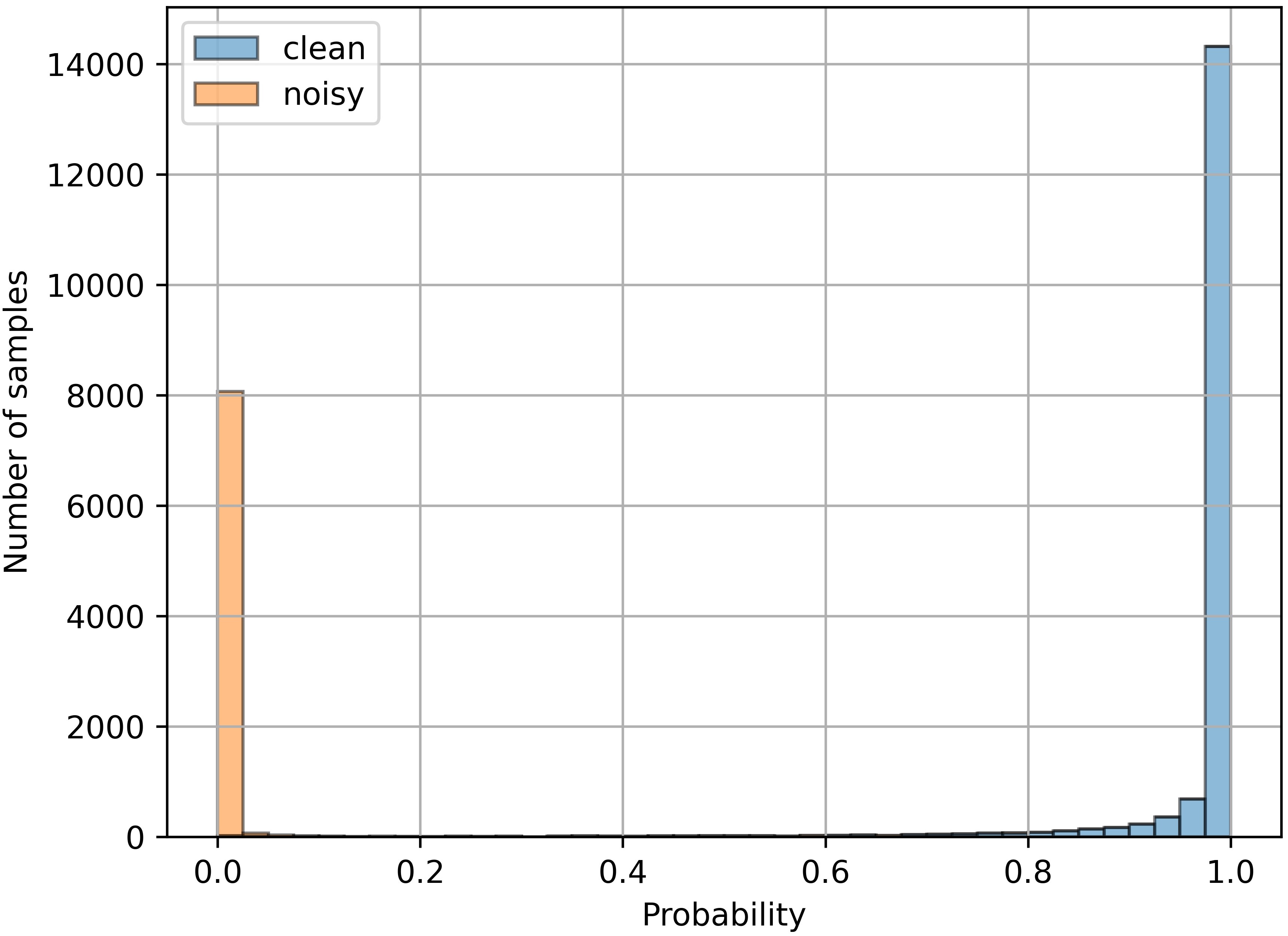}
\caption{Symmetric-noise}
\label{NL_Lim:Symm}
\end{subfigure}
\begin{subfigure}{0.49\linewidth}
\centering
\captionsetup{font=footnotesize}
\includegraphics[width=\linewidth]{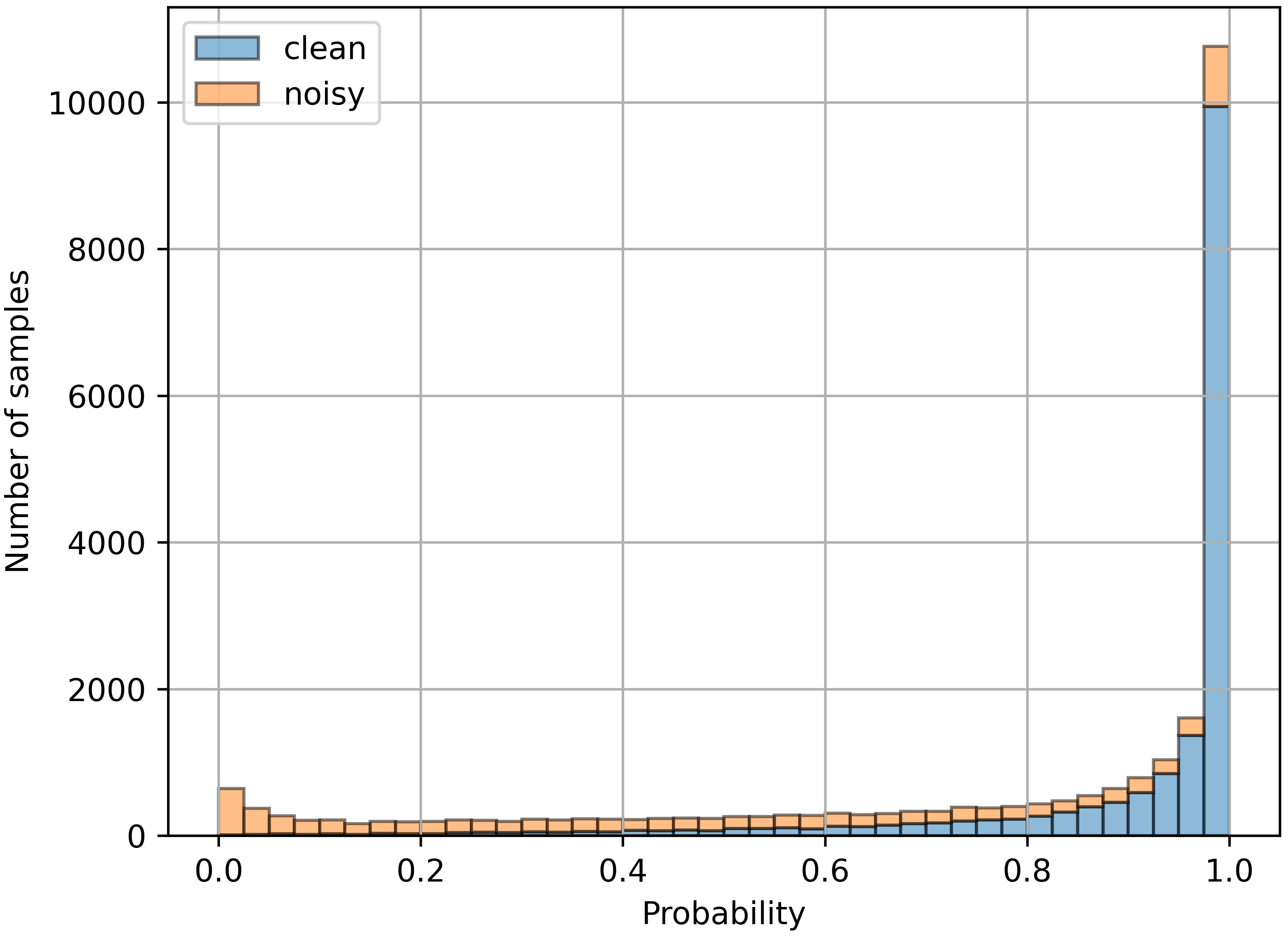}
\caption{Shift-noise}
\label{NL_Lim:Shift}
\end{subfigure}
%
\caption{\small{Histogram showing the noise-filtering performance of 
\cite{kim2019nlnl} on MNIST.
In both cases, the amount of noise equals $32.9\%$ (cf. SVHN$\rightarrow$MNIST shift-noise in Table~\ref{tab:digit5}).
}}
\label{fig:NL_lim_sep}
\end{figure}
\begin{figure*}[!t]
\setlength{\belowcaptionskip}{-5pt}
\captionsetup{font=small}
\centering
\begin{subfigure}{0.24\linewidth}
\captionsetup{font=footnotesize}
\includegraphics[width=\linewidth]{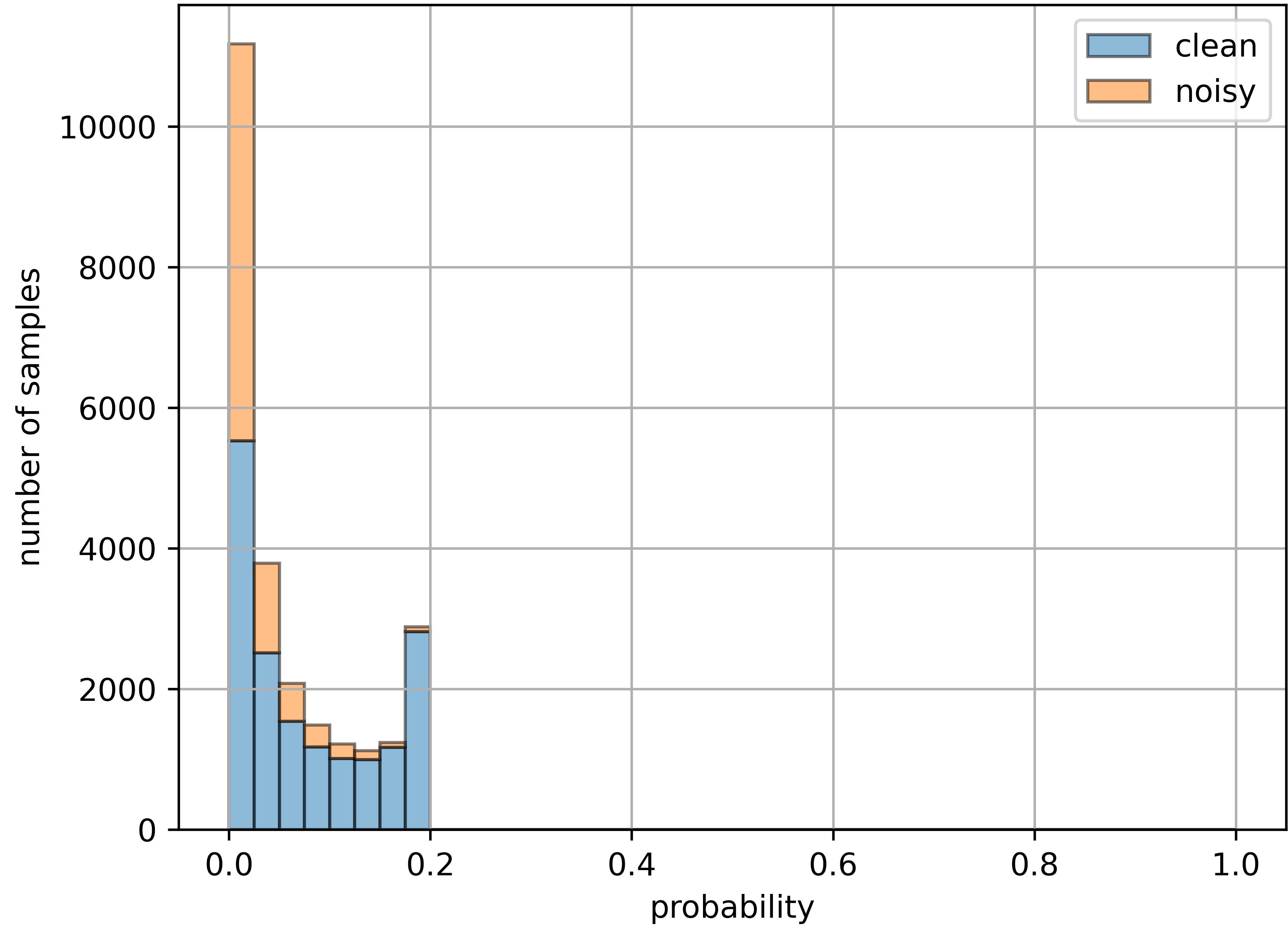}
\caption{epochs = 1}
\label{train:a1_2}
\end{subfigure}
\begin{subfigure}{0.24\linewidth}
\captionsetup{font=footnotesize}
\includegraphics[width=\linewidth]{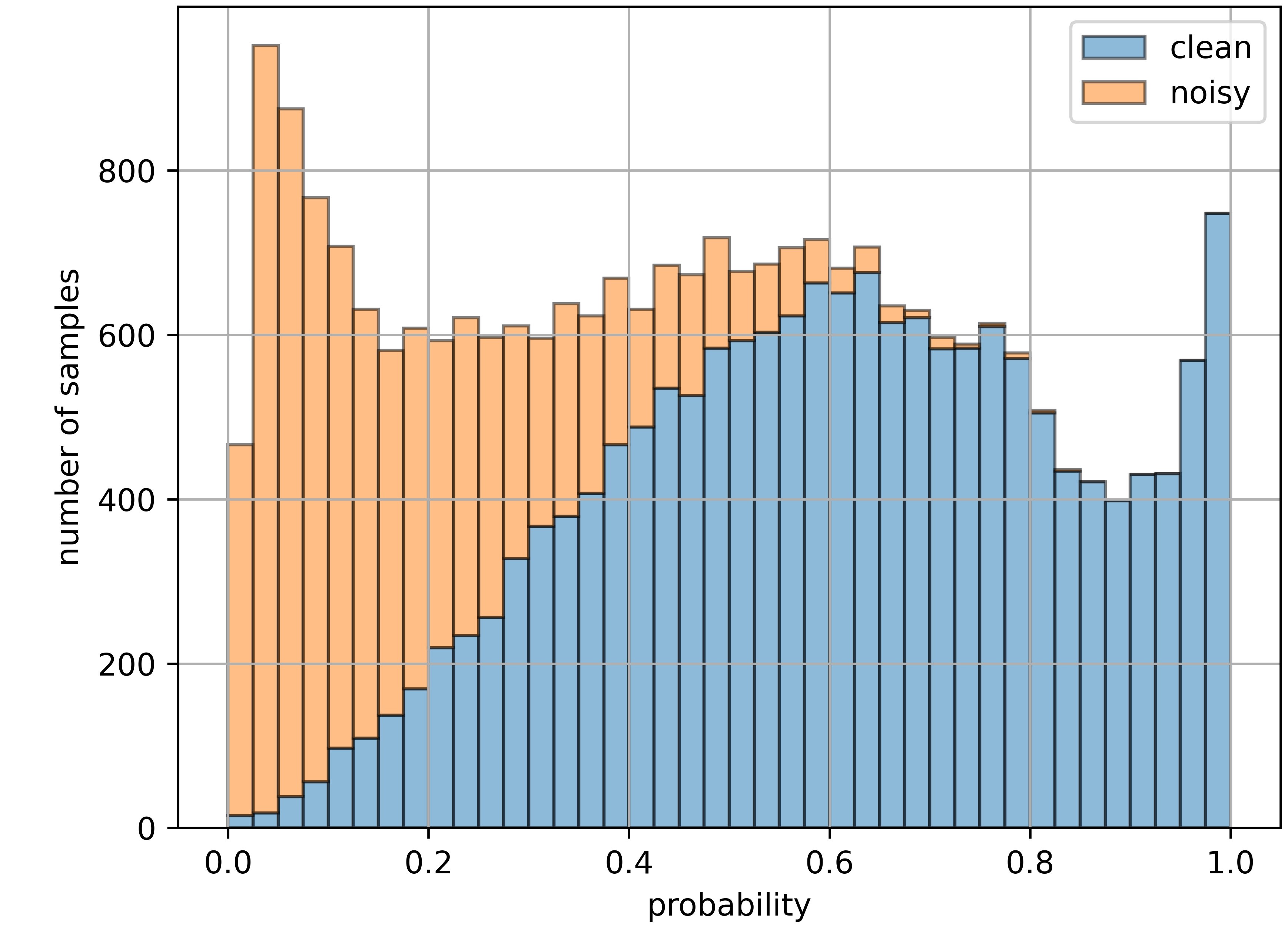}
\caption{epochs = 10}
\label{train:a1_3}
\end{subfigure}
\begin{subfigure}{0.24\linewidth}
\captionsetup{font=footnotesize}
\includegraphics[width=\linewidth]{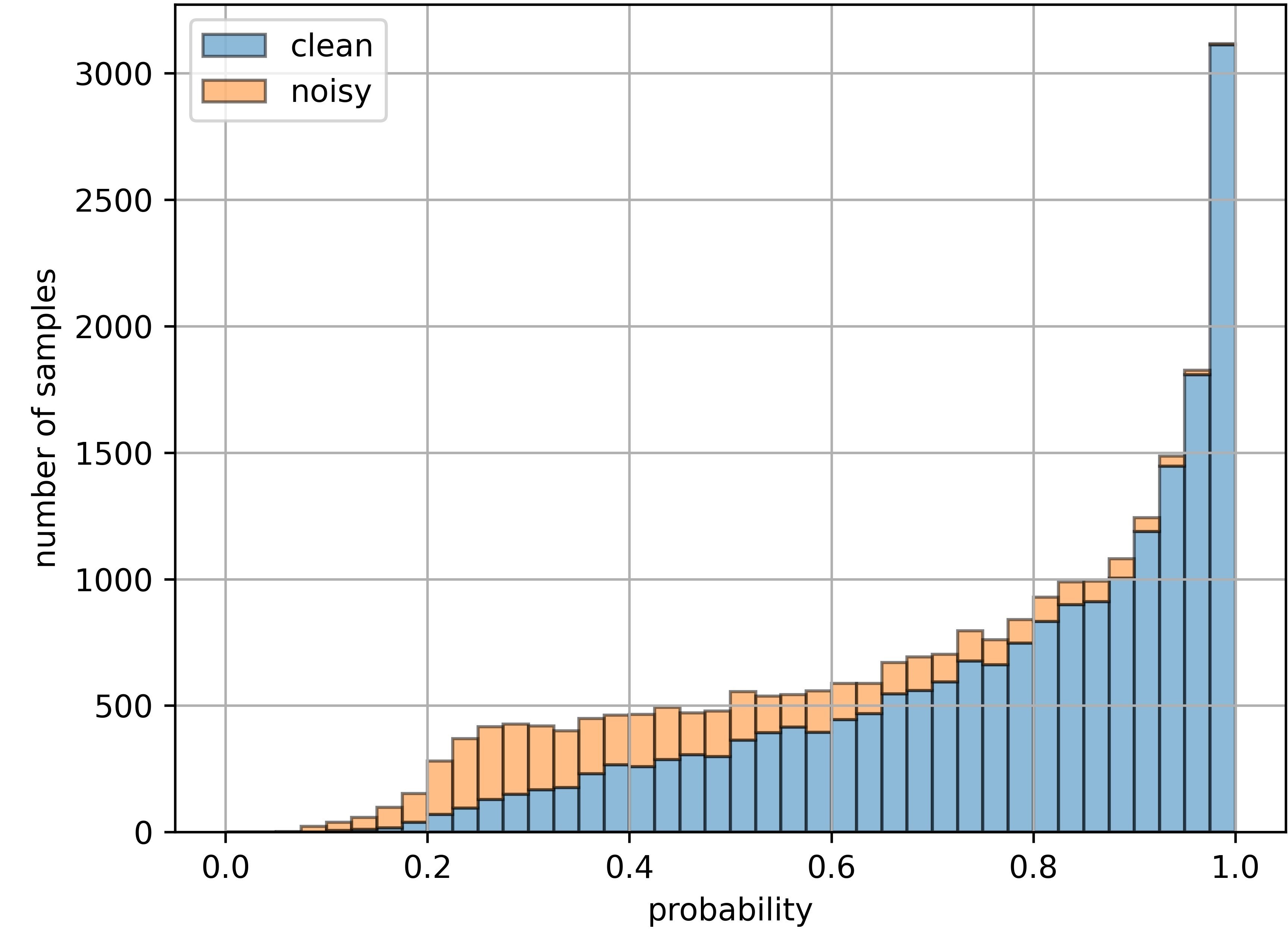}
\caption{epochs = 25}
\label{train:a2_1}
\end{subfigure}
\begin{subfigure}{0.24\linewidth}
\captionsetup{font=footnotesize}
\includegraphics[width=\linewidth]{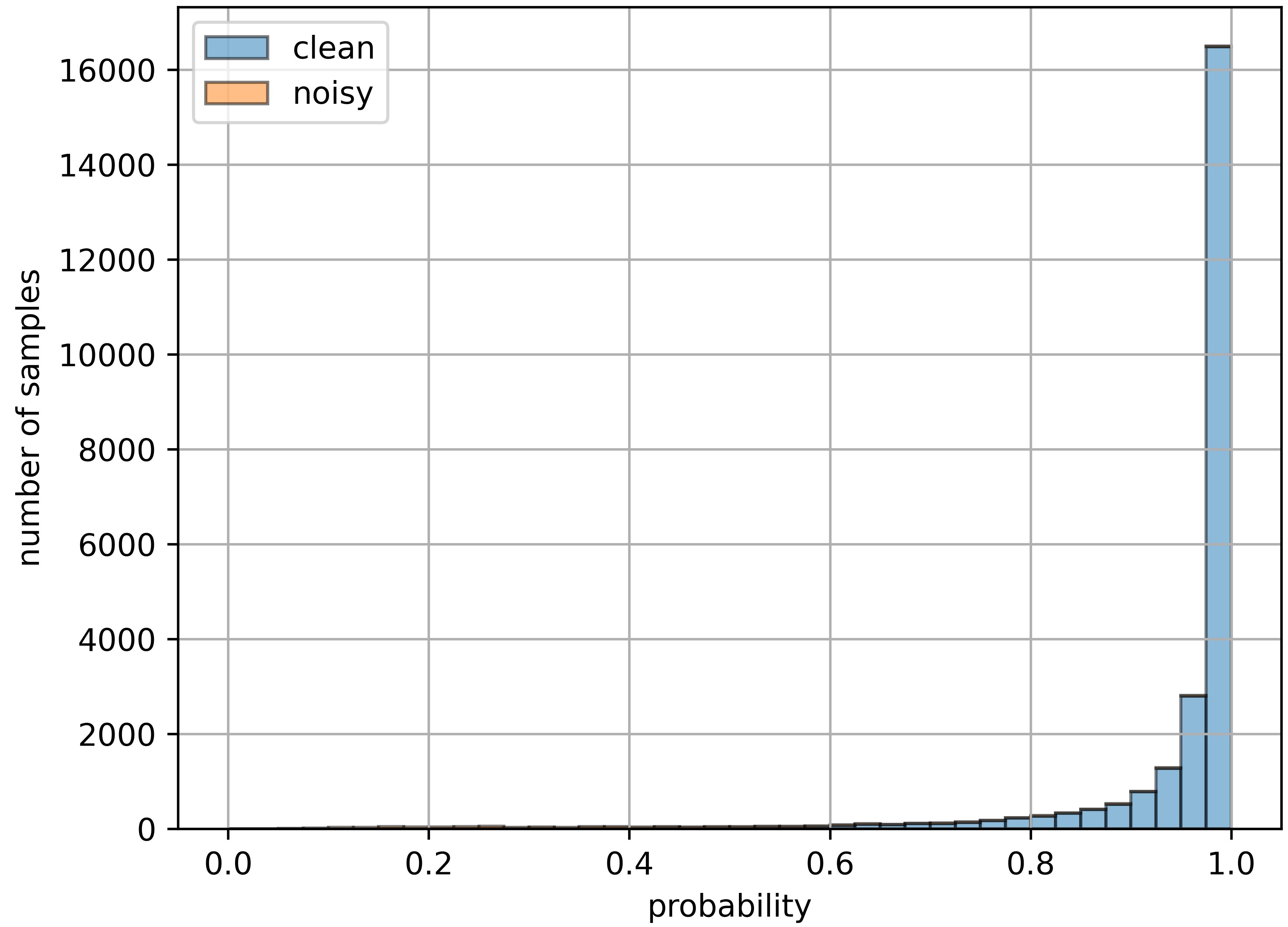}
\caption{epochs = 50}
\label{train:a2_3}
\end{subfigure}
\caption{\small{Prediction-confidence trend during training and  pseudo-label refinement by our proposed AdaPLR method in case of SVHN$\rightarrow$MNIST source-free UDA. Almost all samples are predicted with very low confidence at the beginning (a). As the network starts learning, noisy samples are segregated in a low confidence interval. Only if confidence is lower than $\gamma$ (eq. \ref{eq:gamma}), pseudo-labels are reassigned. Noise is thus progressively reduced (c-d). (Best viewed in color)}}
\label{train:main}
\end{figure*}
%
Nevertheless, the fundamental limitation of naive NL \cite{kim2019nlnl} is that it is suitable for the noise showing a uniform distribution (\textit{symmetric-noise}) only. As reported in Figure~\ref{NL_Lim:Symm}, data samples with labels are assigned low confidence, resulting in effective noise separation.

Yet, a significant amount of noise is overfitted with high confidence by the network when we consider \textit{shift-noise}, namely noise produced by inferring pseudo-labels on the target by using a model trained with the source (MNIST and SVHN, respectively, in the example of Figure~\ref{NL_Lim:Shift}). Note that overfitting is not to be ascribed to the \textit{amount} of noise, which was carefully balanced in the experiment, but rather to its \textit{distribution} (for more details see the Appendix).

\textit{\textbf{Ensemble Learning.}} It refers to concurrently training multiple networks of similar configuration on the same set of inputs. The idea is to create a set of experts trained in different ways, in order to produce predictions with low bias and high variance. 
Generally, the ensemble produces the final output as a weighted sum of all the experts' logits, \ie, for a data sample $\boldsymbol{x}$, the final prediction can be defined as:
\begin{equation}\label{eq:ens}
\boldsymbol{p}_{e} = \sigma(\sum_{k=1}^{N_e}\alpha^k 
\psi^k(\nu^k(\boldsymbol{x}))
\end{equation}

\noindent
where $N_e$ corresponds to the number of experts in the ensemble network and $\alpha^k$ is a set of weights that defines the contribution of each member.
In this work we set $\alpha^k=1, \forall k\in[1,N_e]$ and propose a novel Negative Ensemble Learning technique for learning with noisy labels that, amplifies the diversity of the ensemble members by different stochastic (i) input augmentation, and (ii) feedback using Disjoint Residual Labels. The resulting strong consensus gives rise to the cleaner pseudo-labels, better than those obtained by any stand-alone network.

\subsection{Adaptive Pseudo-Label Refinement (AdaPLR)}
\textbf{Problem Setup.}
The goal of UDA methods is to adapt a model pre-trained on a labelled source domain $\mathcal{D}_s=\{(\boldsymbol{x}^i_s,y^i_s)\}_{i=1}^{N_s}$ in order to generalize well on a different, yet related, unlabeled target domain $\mathcal{D}_t=\{\boldsymbol{x}^j_t\}_{j=1}^{N_t}$, where $N_s$ and $N_t$ denote the number of samples in the source and the target domain, respectively. 
In general, we assume working with $M_d+1$ domains, including a target domain $\mathcal{D}_t$ and source domains $\mathcal{D}_s$, where $s=\{1,...,M_d\}$.
The main distinction between our approach and most standard UDA methods is that we do not use source data nor generate target-style data at any stage of the proposed method. Instead, we use only a pre-trained source model to infer pseudo-labels $\mathcal{P}={\{\tilde{y}^j\}}_{j=1}^{N_t}$ on the target domain. Since this results in a significant amount of incorrect labels (due to domain shift \cite{torralba2011unbiased}), we propose a way to progressively filter out noisy target samples from the clean ones, and carry out pseudo-label refinement to obtain cleaner $\mathcal{P}$.


\textbf{Pseudo-Label Refinement.} The prerequisite step for our proposed method refers to inferring pseudo-labels $\mathcal{P}$ 
of unlabeled data samples of $\mathcal{D}_t$ using $f_s$ -- a pre-trained model on the labeled samples from $\mathcal{D}_s$.
Specifically, for \textit{single-source} UDA, the model pre-trained on the chosen source is used to infer pseudo-labels of every target domain being considered. For \textit{multi-source} UDA, we 
have two possibilities in which off-the-shelf pre-trained models can be used:

$1)$ A single model pre-trained on the dataset of aggregated sources $f_{agg}$ is used to infer $\mathcal{P}$, which is defined as:
\begin{equation}\label{eq:s1a}
\tilde{y}^j = \argmax ( \psi_{agg}( \nu_{agg}(\boldsymbol{x}^j_t)) ) \tab[0.5cm] \forall j \in \{1,...,N_t\},
\end{equation}
%

$2)$ A set of $M_d$ models, each pre-trained on one of the sources considered, $f_1,...,f_{M_d}$, are used to infer $\mathcal{P}$. The adopted strategy in this case is known as late-fusion,
which is defined as,
\vspace{-0.6em}
\begin{equation}\label{eq:s1b}
\tilde{y}^j = \argmax( \sum_{s=1}^{M_d} \psi_s(\nu_s(\boldsymbol{x}_t^j)) )
\tab[0.5cm] \forall j\in \{1,...,N_t\},
\end{equation}

\noindent
In this study, we employ Eq.~(\ref{eq:s1a}) as it represents a more realistic scenario in which a single available model is pre-trained on the maximum data at hand, \ie, data aggregated from all possible sources. Later, for scalability, Eq.~(\ref{eq:s1b}) can be used to leverage additional pre-trained source models.

Subsequently, the proposed AdaPLR method is applied. 
In particular, we obtain robust and cleaner pseudo-labels to refine $\mathcal{P}$ using the moving average of $N_a$ previous ensemble output predictions:

\begin{equation}\label{eq:plr}
\boldsymbol{p} = \sigma ( \frac{1}{N_a \cdot N_e} \sum_{l=1}^{N_{a}}\sum_{k=1}^{N_e} 
\psi^{k,l}_e(\nu^{k,l}_e(\boldsymbol{x}) )
\end{equation}

\noindent
where we set $N_a=10$ for all the experiments in this study.\footnote{We anticipate that this is a non sensitive parameter, it has not a relevant effect on the performance as long as $N_a\geq5$.} Now let us define $\tilde{p}$ as the confidence of the pseudo-label $\tilde{y}$, namely the entry of the probability vector $\boldsymbol{p} $ corresponding to $\tilde{y}$. 
We categorize the target samples with $\tilde{p} > \alpha$ as High Confidence Samples (HCS), and define the ratio of HCS over the total number of samples, as follows:

\begin{equation}\label{eq:gamma}
\gamma = \frac{\# \tab[0.1cm] of \tab[0.1cm] HCS}{N_t}
\end{equation}

During training, the pseudo-label of each sample $\boldsymbol{x}_t^j$ is updated/retained according to the following condition:

\vspace{-1em}
\begin{align}
\tilde{y}^j (n) = \begin{dcases*}
    \argmax(\boldsymbol{p}^j), & if $\tilde{p}^j < \gamma $\\
    \tilde{y}^j (n-1), & otherwise
    \end{dcases*}
    \forall j 
\end{align}

\noindent
where $n$ denotes the epoch number. The intuition behind such reassignment rule can be drawn from the trend of the prediction confidence along training. As shown in Figure~\ref{train:a1_2} and \ref{train:a1_3}, with the growing number of epochs, noisy samples remain towards low confidence regime and clean samples obtain high confidence progressively. In Figure~\ref{train:a1_3}, the ratio of HCS (with $\alpha=0.9$) gives $\gamma\approx0.15$ that corresponds to a confidence region $[0,\gamma] = [0,0.15]$ in which the noisy samples are prevalent, hence they are the best candidates to be subjected to 
label reassignment. Consequently, pseudo-label refinement is achieved progressively during training in an adaptive manner (see Figure~\ref{train:a2_1} and \ref{train:a2_3}, where total noise is progressively reduced). We ablate to find optimal value of $\alpha$ in Section \ref{AblationStudy}.

\subsection{Negative Ensemble learning} Working with an ensemble allows us to make a smart use of more than one complementary label, \ie, \textit{Residual Labels (RL)}, for each sample. So, let us consider the case we pick $N_{RL}$ Residual Labels at once: this influences the training process in three ways:
(a) The likelihood of the actual-label $y_t$ being randomly picked as one of the complementary labels is now $\frac{N_{RL}}{C - 1}$, which is bad. 
%
(b) In case $y_t\cap RL=\varnothing$, the training is accelerated by the stronger feedback provided by the multiple contributions of RLs, which is good.
(c) In case $y_t\cap RL\neq\varnothing$, instead of providing entirely wrong feedback using Eq.~\ref{eq:nl}, the impact of wrong feedback is mitigated by a factor of $\frac{N_{RL}-1}{N_{RL}}$, and this is again good. In fact, gradients will follow a mean direction according to the new Negative Ensemble Learning (NEL) loss that we define as:
\vspace{-0.5em}
\begin{equation}\label{eq:nl-RL}
\mathcal{L}_{\small{NEL}}(\mathcal{D}_t)=-\mathbb{E}_{x_t \sim\mathcal{D}_t} \frac{1}{N_{RL}}\sum_{c=1}^{C} \mathds{1}_{[c \in RL]}log(1-\boldsymbol{p}_c)
\end{equation}

Although (b) \& (c) above are essential advantages of using multiple RL, (a) is a downside. Therefore, we ablate on the different values of $N_{RL}$ in Section~\ref{AblationStudy}.

\begin{figure}[!t]
\setlength{\belowcaptionskip}{-5pt}
\centering
\includegraphics[width=0.9\linewidth,height=5cm]{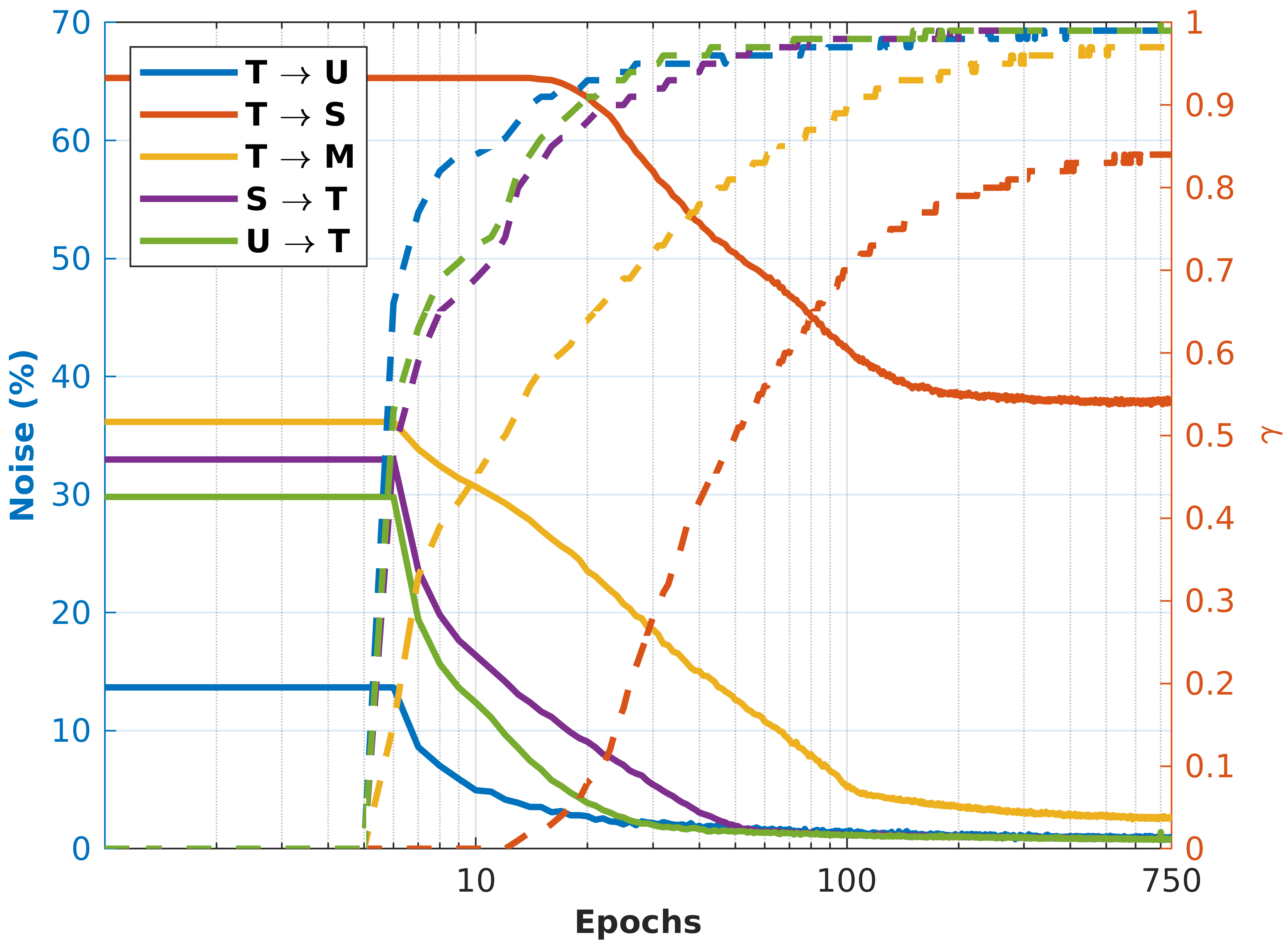}
\vspace{-0.3em}
\caption{\small{
Correlation between adaptiveness of $\gamma$ threshold (right $y$-axis, dashed profiles) and progressive noise reduction (left $y$-axis, solid profiles) achieved by AdaPLR during training for various amount of noise. 
Legend: \textit{\textcolor{blue}{\textbf{T}}: MNIST, \textcolor{blue}{\textbf{S}}: SVHN, \textcolor{blue}{\textbf{U}}: USPS, and \textcolor{blue}{\textbf{M}}: MNIST-M.}}}
\label{fig:lambda}
\end{figure}

\textbf{Disjoint Residual Labels.} We further extend the idea of RL by introducing the concept of Disjoint Residual Labels (DRL) which is especially suited for ensembles, ($N_e>1$). Each ensemble member can in fact be trained using Eq. \ref{eq:nl-RL}, but with a completely \textit{disjoint} stochastic subset of residual labels. Not only this allows each member to receive a different feedback (thus enhancing the ensemble's diversity) but also restricts the possibility of receiving wrong feedback to one member only.

\textbf{Stochastic Augmentations.} 
The second strategy to induce additional diversity in ensemble members consists in feeding each member with the same batch of input images
with different stochastic data augmentation. We consider several stochastic data augmentation strategies including the composition of (i) spatial/geometric transformation via random cropping (with uniform area = $0.08$ to $1.0$ and aspect-ratio = $\frac{3}{4}$ to $\frac{4}{3}$) followed by resizing to the original size, (ii) affine transformation followed by Gaussian blur, and (iii) color distortion.
We found that composition of different stochastic data augmentation is crucial to avoid noise overfitting and extend diversity in the ensemble network (See Section~\ref{AblationStudy}).

\begin{figure}[t]
\setlength{\belowcaptionskip}{-5pt}
\captionsetup{font=small}
\centering
\includegraphics[width=0.95\linewidth]{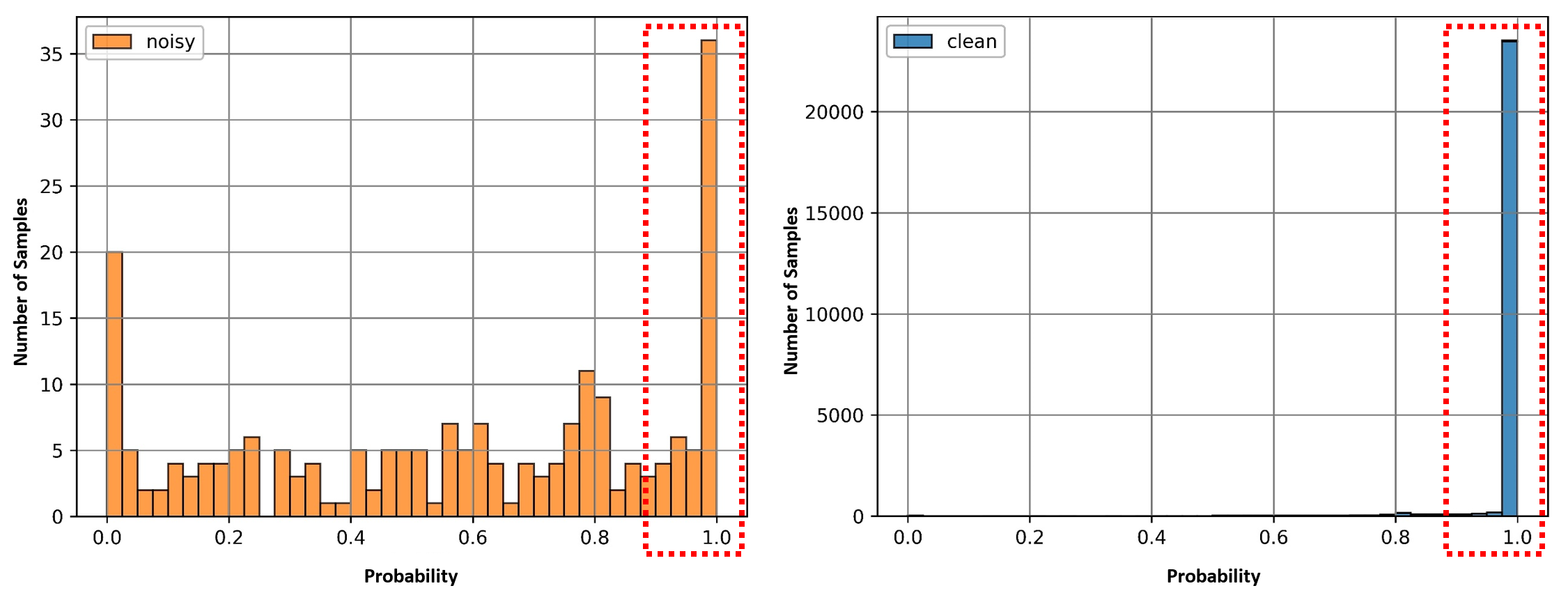}
\caption{Distribution of remaining noise in refined pseudo-labels after SVHN$\rightarrow$MNIST UDA. The highlighted bars represent the set of samples with confidence greater than  $\alpha=0.9$.}
\label{fig:FPL}
\end{figure}

All these ingredients concur in solving the shortcomings highlighted in Fig. \ref{fig:NL_lim_sep}. AdaPLR is in fact capable of filtering out different amounts of noise in an adaptive manner while refining pseudo-labels progressively. Trends with respect to right $y$-axis in Figure~\ref{fig:lambda} show the capability of the threshold $\gamma$ to adapt to different single-source UDA cases, while trends with respect to left $y$-axis show the corresponding noise reduction.
In particular, it is worth noting that no pseudo-label refinement is applied during the first few epochs, until the ensemble gets a mature state of generalization capability.

After a certain number of training epochs, the pseudo-label refinement process stalls down to an insignificant noise reduction rate (see Figure~\ref{fig:lambda}). 
Therefore, instead of pushing AdaPLR more for several epochs to achieve further small reduction, we apply standard supervised learning only using high confidence samples determined by $\alpha$. This is done using only one model (the final target model) for a fair comparison with state-of-the-art methods. As shown in Figure~\ref{fig:FPL}, the remaining noise is distributed over the entire probability spectrum, whereas the majority of the clean samples are predicted with high confidence (red box). 
Therefore, only a small fraction of noisy samples affects training.

\begin{table*}
\centering
\captionsetup{width=.85\textwidth,font=small}
\footnotesize
\begin{tabular}{|l|c c c c c |c||l|c c c c c |c|}
\hline
& \multicolumn{6}{c||}{\textbf{Single-Source UDA}} & \multicolumn{7}{c|}{\textbf{Multi-Source UDA}} \\
\hline
\multirow{2}{*}{Source} & 
\multirow{2}{*}{\textcolor{blue}{\textbf{\textit{T}}}} & 
\multirow{2}{*}{\textcolor{blue}{\textbf{\textit{T}}}} & 
\multirow{2}{*}{\textcolor{blue}{\textbf{\textit{T}}}} & 
\multirow{2}{*}{\textcolor{blue}{\textbf{\textit{S}}}} &
\multicolumn{1}{ c| }{\multirow{2}{*}{\textcolor{blue}{\textbf{\textit{U}}}}} & 
\multirow{3}{*}{Avg.} & 
&
\textcolor{blue}{\textbf{\textit{M,S,}}} & 
\textcolor{blue}{\textbf{\textit{T,S,}}} &
\textcolor{blue}{\textbf{\textit{T,M,}}} &
\textcolor{blue}{\textbf{\textit{T,M,}}} &
\multicolumn{1}{ c| }{\textcolor{blue}{\textbf{\textit{T,M,}}}} & 
\multirow{3}{*}{Avg.} 
\\
 & & & & & \multicolumn{1}{ c| }{ } & & 
&
\textcolor{blue}{\textbf{\textit{D,U}}} & 
\textcolor{blue}{\textbf{\textit{D,U}}} &
\textcolor{blue}{\textbf{\textit{D,U}}} &
\textcolor{blue}{\textbf{\textit{S,U}}} &
\multicolumn{1}{ c| }{\textcolor{blue}{\textbf{\textit{S,D}}}} & 
 
\\
\cline{1-6}
\cline{8-13}
Target & 
\textcolor{blue}{\textbf{\textit{U}}} &
\textcolor{blue}{\textbf{\textit{S}}} &
\textcolor{blue}{\textbf{\textit{M}}} &
\textcolor{blue}{\textbf{\textit{T}}} &
\multicolumn{1}{ c| }{\textcolor{blue}{\textbf{\textit{T}}}} &
&
&
\textcolor{blue}{\textbf{\textit{T}}} &
\textcolor{blue}{\textbf{\textit{M}}} &
\textcolor{blue}{\textbf{\textit{S}}} &
\textcolor{blue}{\textbf{\textit{D}}} &
\multicolumn{1}{ c| }{\textcolor{blue}{\textbf{\textit{U}}}} & 
\\
\hline
ATT \cite{pmlr-v70-saito17a} & -- & 52.8 & 94.0 & 85.8 & -- & -- & 
DCTN \cite{xu2018deep} & -- & 70.9 & 77.5 & -- & -- & -- \\
SBA \cite{russo2018source} & 97.1 & 50.9 & \textbf{98.4} & 74.2 & 87.5 & 81.6 &
MM \cite{peng2019moment} & 98.4 & 72.8 & 81.3 & 89.5 & 96.1 & 87.6 \\
MALT \cite{minnehan2019deep} & 97.0 & \textbf{78.7} & 71.4 & 98.7 & 20.7 & 73.3 &
OML \cite{li2020online} & 98.7 & 71.7 & 84.8 & 91.1 & 97.8 & 88.8 \\
MTDA \cite{gholami2020unsupervised} & 94.2 & 52.0 & 85.5 & 84.6 & 91.5 & 81.5 &
CMSS \cite{yang2020curriculum} & 99.0 & 75.3 & 88.4 & \textbf{93.7} & 97.7 & 90.8 \\
GPLR \cite{morerio2020generative} & 89.3 & 63.4 & 94.3 & 97.3 & 91.8 & 87.5 &
& & & & & & \\
\hline
AdaPLR & \textbf{97.4} & 61.6 & 95.4 & \textbf{99.2} & \textbf{99.2} & \textbf{90.6} &
& \textbf{99.1} & \textbf{95.5} & \textbf{89.6} & 90.0 & \textbf{97.8} & \textbf{94.4} \\
\hline
\end{tabular}
\vspace{-0.9em}
\caption{\small{Classification accuracy on Digit5 with a naive 3-layer CNN. 
Legend: \textit{\textcolor{blue}{\textbf{T}}: MNIST, \textcolor{blue}{\textbf{S}}: SVHN, \textcolor{blue}{\textbf{U}}: USPS, \textcolor{blue}{\textbf{M}}: MNIST-M, and \textcolor{blue}{\textbf{D}}: Synthetic-Digits.}}}
\vspace{-0.5em}
\label{tab:digit5}
\end{table*}
\begin{table*}[ht]
\centering
\captionsetup{width=.85\textwidth,font=small}
\footnotesize
\begin{tabular}{|l|c c c c c c |c||l|c c c c |c|}
\hline
& \multicolumn{7}{c||}{\textbf{Multi-Target UDA}} & \multicolumn{6}{c|}{\textbf{Multi-Source UDA}} \\
\hline
Source & 
\textcolor{blue}{\textbf{\textit{ }}} & 
\textcolor{blue}{\textbf{\textit{P}}} & 
\multicolumn{1}{ c| }{\textcolor{blue}{\textbf{\textit{ }}}} & 
\textcolor{blue}{\textbf{\textit{ }}} &
\textcolor{blue}{\textbf{\textit{A}}} &
\multicolumn{1}{ c| }{\textcolor{blue}{\textbf{\textit{ }}}} & 
\multirow{2}{*}{Avg.} & 
&
\textcolor{blue}{\textbf{\textit{C,P,S}}} & 
\textcolor{blue}{\textbf{\textit{A,P,S}}} &
\textcolor{blue}{\textbf{\textit{A,C,S}}} &
\multicolumn{1}{ c| }{\textcolor{blue}{\textcolor{blue}{\textbf{\textit{A,C,P}}}}} & 
\multirow{2}{*}{Avg.} 
\\
\cline{1-7}
\cline{9-13}
Target & 
\textcolor{blue}{\textbf{\textit{A}}} &
\textcolor{blue}{\textbf{\textit{C}}} &
\multicolumn{1}{ c| }{\textcolor{blue}{\textbf{\textit{S}}}} & 
\textcolor{blue}{\textbf{\textit{P}}} &
\textcolor{blue}{\textbf{\textit{C}}} &
\multicolumn{1}{ c| }{\textcolor{blue}{\textcolor{blue}{\textbf{\textit{S}}}}} &
&
&
\textcolor{blue}{\textbf{\textit{A}}} &
\textcolor{blue}{\textbf{\textit{C}}} &
\textcolor{blue}{\textbf{\textit{P}}} &
\multicolumn{1}{ c| }{\textcolor{blue}{\textcolor{blue}{\textbf{\textit{S}}}}} & 
\\
\hline
1-NN* & 15.2 & 18.1 & 25.6 & 22.7 & 19.7 & 22.7 & 20.7 & 
DD \cite{mancini2018boosting}       & 87.5 & 87.0 & 96.6 & 71.6 & 85.7 \\
ADDA* & 24.3 & 20.1 & 22.4 & 32.5 & 17.6 & 18.9 & 22.6 &
SIB \cite{Hu2020Empirical}          & 88.9 & 89.0 & 98.3 & 82.2 & 89.6 \\
DSN* & 28.4 & 21.1 & 25.6 & 29.5 & 25.8 & 24.6 & 25.8 &
OML \cite{li2020online}            & 87.4 & 86.1 & 97.1 & 78.2 & 87.2 \\
ITA* & 31.4 & 23.0 & 28.2 & 35.7 & 27.0 & 28.9 & 29.0 &
RABN \cite{xu2019self}              & 86.8 & 86.5 & 98.0 & 71.5 & 85.7 \\
KD \cite{belal2021knowledge} & 24.6 & 32.2 & \textbf{33.8} & 35.6 & 46.6 & \textbf{57.5} & 46.6 & 
JiGen \cite{carlucci2019domain}     & 84.8 & 81.0 & 97.9 & 79.0 & 85.7 \\
& & & & & & & & 
CMSS \cite{yang2020curriculum}      & 88.6 & \textbf{90.4} & 96.9 & 82.0 & 89.5 \\
\hline
AdaPLR & \textbf{80.1} & \textbf{76.1} & 25.9 & \textbf{96.0} & \textbf{82.8} & 49.8 & \textbf{68.4} & 
& \textbf{90.8} & 89.5 & \textbf{98.8} & \textbf{85.2} & \textbf{91.1} \\
\hline
\end{tabular}
\vspace{-0.9em}
\caption{\small{Classification accuracy on PACS with ResNet18.
* results are taken from \cite{gholami2020unsupervised}.
Legend:  \textit{\textcolor{blue}{\textbf{A}}: Art-Painting, \textcolor{blue}{\textbf{C}}: Cartoon, \textcolor{blue}{\textbf{P}}: Photo, and \textcolor{blue}{\textbf{S}}: Sketch.}} 
}
\vspace{-2em}
\label{tab:PACS}
\end{table*}
\begin{table}[!t]
\vspace{0.5em}
\centering
\captionsetup{width=.97\linewidth,font=small}
\footnotesize
\resizebox{\linewidth}{!}{
\begin{tabular}{|l|c c c c c c |c|}
\hline
& \multicolumn{7}{c|}{\textbf{Single-Source UDA}} \\
\hline
Source & 
\textcolor{blue}{\textbf{\textit{P}}} &
\textcolor{blue}{\textbf{\textit{P}}} &
\textcolor{blue}{\textbf{\textit{P}}} &
\textcolor{blue}{\textbf{\textit{A}}} &
\textcolor{blue}{\textbf{\textit{A}}} &
\textcolor{blue}{\textbf{\textit{A}}} &
\multirow{2}{*}{Avg.} 
\\
\cline{1-7}
Target & 
\textcolor{blue}{\textbf{\textit{A}}} &
\textcolor{blue}{\textbf{\textit{C}}} &
\textcolor{blue}{\textbf{\textit{S}}} &
\textcolor{blue}{\textbf{\textit{P}}} &
\textcolor{blue}{\textbf{\textit{C}}} &
\textcolor{blue}{\textbf{\textit{S}}} &
\\
\hline
AdaPLR & \textbf{82.6} & \textbf{80.5} & \textbf{32.3} & \textbf{98.4} & \textbf{84.3} & \textbf{56.1} & \textbf{72.4} \\
\hline
\end{tabular}
}
\vspace{-0.9em}
\caption{\small{Classification accuracy on PACS with ResNet18.}
}
\vspace{-0.9em}
\label{tab:PACS_ssda}
\end{table}
\begin{table*}[ht]
\centering
\captionsetup{font=small}
\footnotesize
\begin{tabular}{|l|c c c c c c c c c c c c |c|}
\hline
Methods & \textcolor{blue}{\textbf{\textit{plane}}} & \textcolor{blue}{\textbf{\textit{bcycl}}} & \textcolor{blue}{\textbf{\textit{bus}}} & \textcolor{blue}{\textbf{\textit{car}}} & \textcolor{blue}{\textbf{\textit{horse}}} & \textcolor{blue}{\textbf{\textit{knife}}} & \textcolor{blue}{\textbf{\textit{mcycl}}} & \textcolor{blue}{\textbf{\textit{person}}} & \textcolor{blue}{\textbf{\textit{plant}}} & \textcolor{blue}{\textbf{\textit{skate}}} & \textcolor{blue}{\textbf{\textit{train}}} & \textcolor{blue}{\textbf{\textit{truck}}} & Avg.  \\
\hline
MCD \cite{saito2018maximum} & 87.0 & 60.9 & 83.7 & 64.0 & 88.9 & 79.6 & 84.7 & 76.9 & 88.6 & 40.3 & 83.0 & 25.8 & 71.9 \\
GPDA \cite{kim2019unsupervised} & 83.0 & 74.3 & 80.4 & 66.0 & 87.6 & 75.3 & 83.8 & 73.1 & 90.1 & 57.3 & 80.2 & 37.9 & 73.3 \\
SAFN \cite{xu2019larger} & 93.6 & 61.3 & 84.1 & 70.6 & 94.1 & 79.0 & \textbf{91.8} & 79.6 & 89.9 & 55.6 & 89.0 & 24.4 & 76.1 \\
DSBN \cite{chang2019domain} & \textbf{94.7} & \textbf{86.7} & 76.0 & 72.0 & \textbf{95.2} & 75.1 & 87.9 & 81.3 & \textbf{91.1} & 68.9 & 88.3 & 45.5 & 80.2 \\
DADA \cite{tang2020discriminative} & 92.9 & 74.2 & 82.5 & 65.0 & 90.9 & \textbf{93.8} & 87.2 & 74.2 & 89.9 & 71.5 & 86.5 & 48.7 & 79.8 \\
\hline
AdaPLR & 94.5 & 60.8 & \textbf{92.3} & \textbf{87.3} & 87.3 & 93.2 & 87.6 & \textbf{91.1} & 56.9 & \textbf{83.4} & \textbf{93.7} & \textbf{86.6} & \textbf{84.2} \\
\hline
\end{tabular}
\vspace{-0.9em}
\caption{\small{Classification accuracy on Visda-C with ResNet101.}}
\vspace{-1.5em}
\label{tab:VisdaC}
\end{table*}
\begin{table}[!ht]
\centering
\captionsetup{width=.97\linewidth,font=small}
\footnotesize
\resizebox{\linewidth}{!}{
\begin{tabular}{|l|c c c c c c |c|}
\hline
Target & 
\textcolor{blue}{\textbf{\textit{C}}} &
\textcolor{blue}{\textbf{\textit{I}}} &
\textcolor{blue}{\textbf{\textit{P}}} &
\textcolor{blue}{\textbf{\textit{Q}}} &
\textcolor{blue}{\textbf{\textit{R}}} &
\textcolor{blue}{\textbf{\textit{S}}} &
Avg.
\\
\hline
MM \cite{peng2019moment} & 58.6 & 26.0 & 52.3 & 6.3 & 62.7 & 49.5 & 42.6 \\
OML \cite{li2020online}  & 62.8 & 21.3 & 50.5 & 15.4 & 64.5 & 50.4 & 44.1 \\
CMSS \cite{yang2020curriculum} & 64.2 & \textbf{28.0} & 53.6 & 16.0 & 63.4 & 53.8 & 46.5 \\
\hline
AdaPLR & \textbf{68.3} & 22.1 & \textbf{54.7} & \textbf{22.8} & \textbf{67.3} & \textbf{57.1} & \textbf{48.7} \\
\hline
\end{tabular}
}
\vspace{-0.9em}
\caption{\small{Classification accuracy on DomainNet with ResNet101. For each target, the rest of the domains are considered as source (multi-source UDA).
Legend: \textit{\textcolor{blue}{\textbf{C}}: Clipart, \textcolor{blue}{\textbf{I}}: Infograph, \textcolor{blue}{\textbf{P}}: Painting, \textcolor{blue}{\textbf{Q}}: Quickdraw, \textcolor{blue}{\textbf{R}}: Real, and \textcolor{blue}{\textbf{S}}: Sketch.}}
}
\vspace{-0.9em}
\label{tab:domainnet}
\end{table}

\section{Experiments}\label{EXP}

\subsection{Datasets and implementation details}\label{ImplDetails}
We consider the image classification task to comprehensively evaluate the proposed AdaPLR method on major UDA benchmarks including Digit5 (MNIST \cite{lecun1998gradient}, SVHN \cite{netzer2011reading}, USPS \cite{denker1989neural}, MNIST-M \cite{ganin2015unsupervised}, and Synthetic-Digits \cite{ganin2015unsupervised}), PACS \cite{li2017deeper}, VisDA-C \cite{peng2018visda}, and DomainNet \cite{peng2019moment}. 
For all the benchmarks, we use batch-size $= 32$, $\alpha=0.9$ 
and Adam as the optimizer with a weight decay of $5e^{-4}$. The base learning rate is set to $1e^{-4}$ and the feature extractors are 
optimized with a learning rate of $1e^{-5}$.  
The feature extractor of ensemble members are initialized with a pre-trained source model. 
For single-source and multi-target UDA, we consider one pre-trained source model to adapt on every target domain. For multi-source UDA, each domain is selected as the target domain while the rest of the domains are treated as aggregated source (according to Eq.~\ref{eq:s1a}).

\textbf{Digit5} refers to a set of digit benchmarks. 
In this paper, following \cite{peng2019moment}, we sample a subset of 25000 images from the training and 9000 images from the testing set for MNIST, MINST-M, SVHN, and Synthetic-Digits.
Since USPS contain a total of 9298 images, we use standard train-test splits.
To keep comparable image resolution, we resize all images to 32×32 and a naive 3-layer CNN is used as ensemble members.
For single-source and multi-source UDA, label refinement takes 750 and 300 epochs, respectively, whereas the final target model is trained for 200 epochs.

\textbf{PACS} 
contains 4 domains, namely \textit{(Art-Painting, Cartoon, Photo, and Sketch)}. There are only 9991 images of 227x227 resolution from 7 object categories that accommodates a large domain shift due to the different image style depictions. We use ResNet-18 as ensemble members. For single-source, multi-target and multi-source UDA, label refinement takes 200, 200 and 100 epochs, respectively, whereas the final target model is trained for 200 epochs.

\textbf{Visda-C} is a challenging large-scale benchmark attempting to bridge the significant synthetic-to-real domain gap across 12 object categories. We follow standard protocol in which the source domain (training split) contains 152K synthetic images 
and the target domain (testing split) contains 72K real images.
We resize all images to 256×256 resolution and use ResNet-101 as ensemble members. 
Label refinement takes 150 epochs and just 25 epochs were found to be enough for training the final target model.

\textbf{DomainNet} is by far the largest UDA benchmark with 6 domains, 600K images and 345 categories. We resize all images to 256×256 and use ResNet-101 as ensemble members. It was mainly developed for multi-source UDA task for which our proposed label refinement takes 100 epochs for \textit{Infogragh} and \textit{Quickdraw} domain, while 40 epoch were enough for the rest. For training the final target model, 100 epochs were sufficient in all the cases.

\subsection{Ablation Study}\label{AblationStudy}
\begin{figure}[!h]
\captionsetup{font=small}
\centering
\begin{subfigure}{.49\linewidth}
\captionsetup{font=footnotesize}
\includegraphics[width=\linewidth]{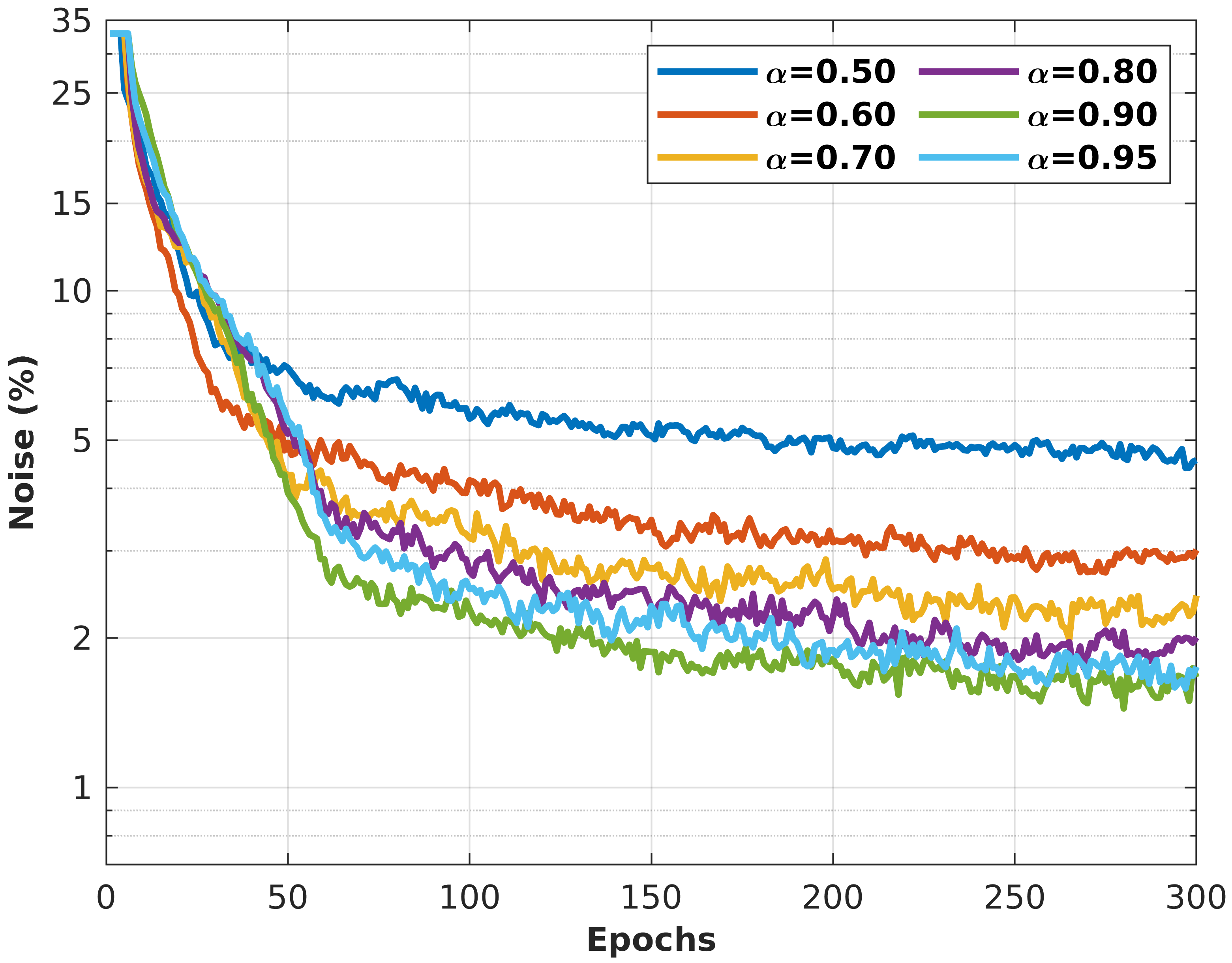} 
\caption{}
\label{ablate:sub-1}
\end{subfigure}
\begin{subfigure}{.49\linewidth}
\captionsetup{font=footnotesize}
\includegraphics[width=\linewidth,height=3.15cm]{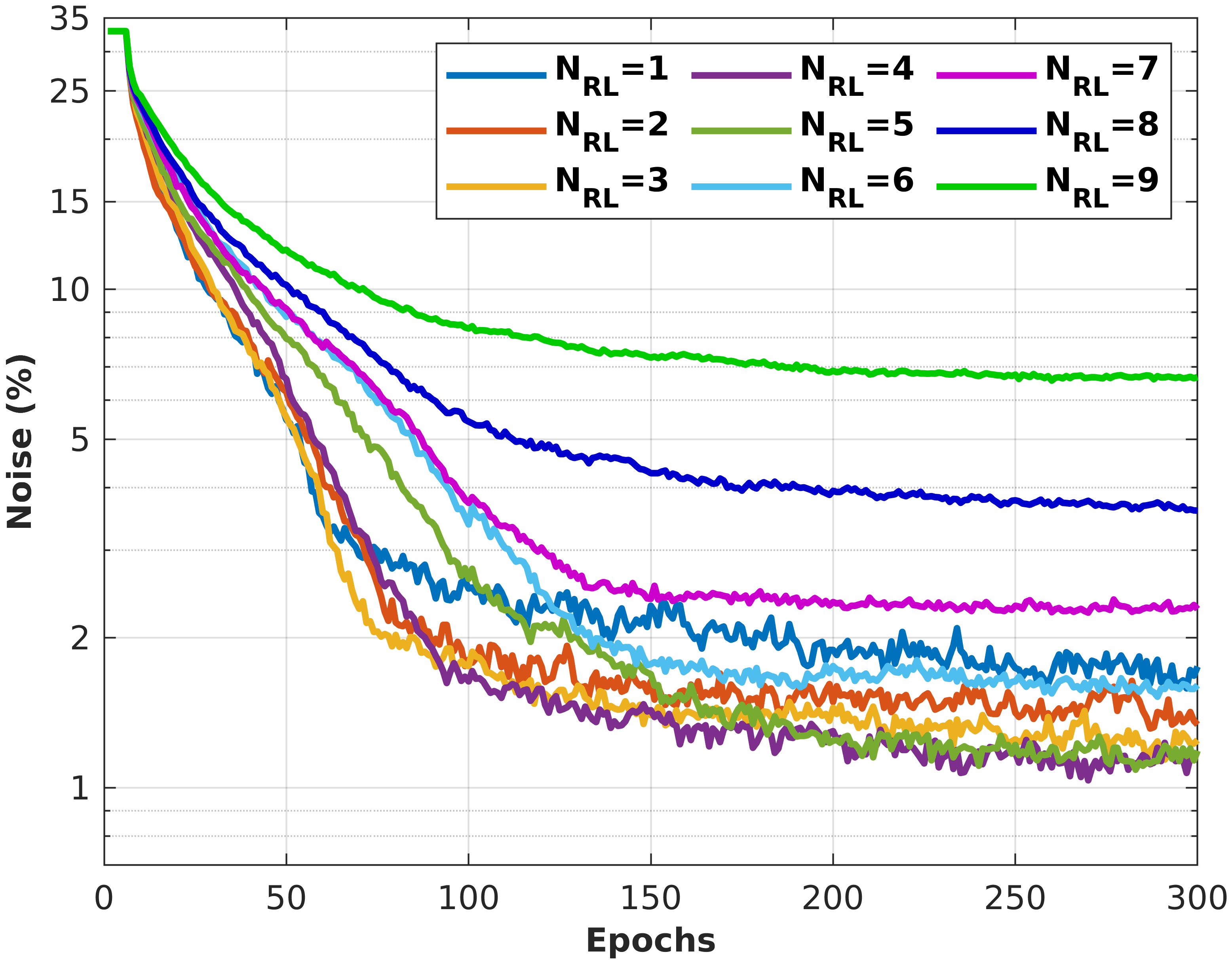}
\caption{}
\label{ablate:sub-2}
\end{subfigure}
\begin{subfigure}{.49\linewidth}
\captionsetup{font=footnotesize}
\includegraphics[width=\linewidth]{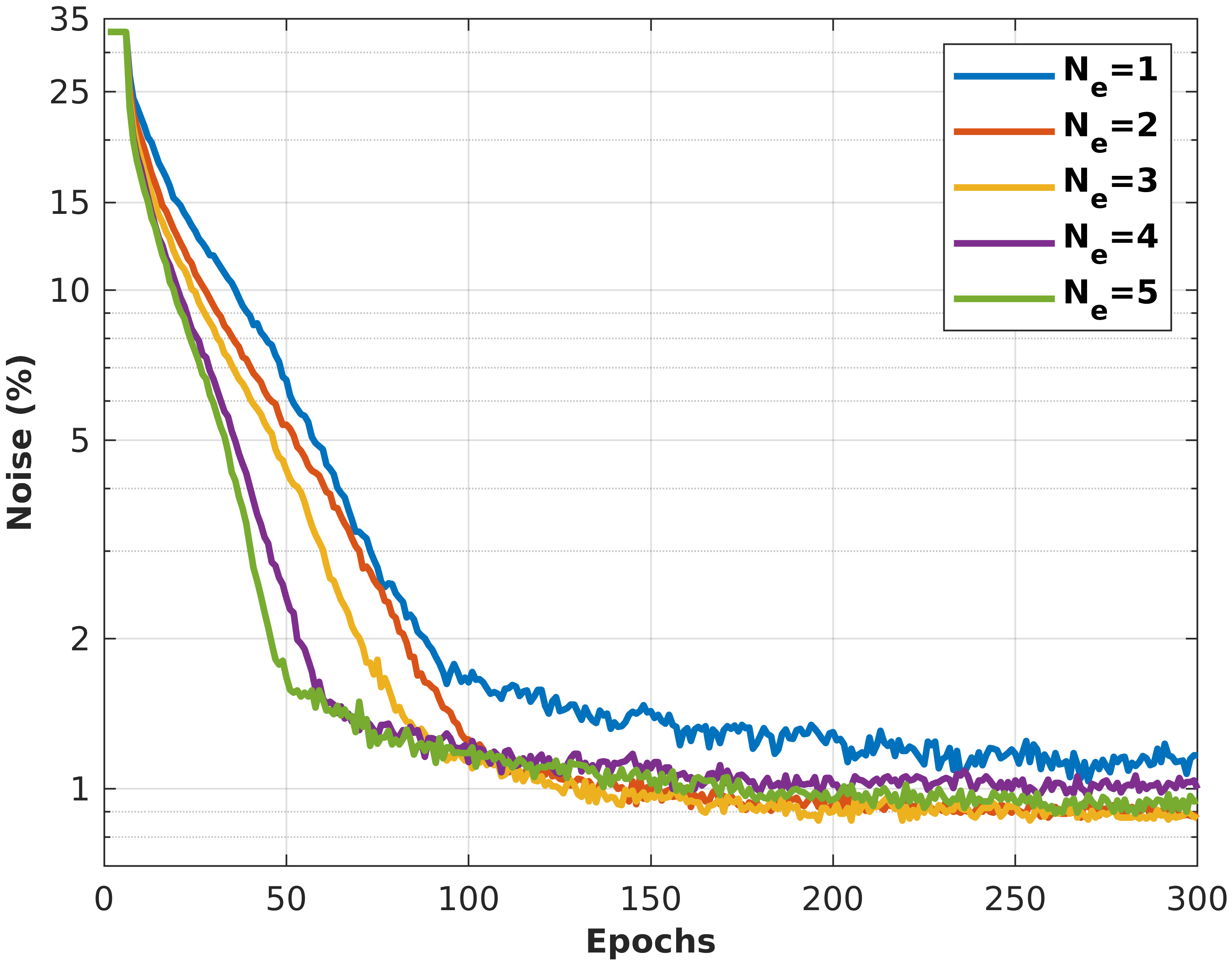}
\caption{}
\label{ablate:sub-3}
\end{subfigure}
\begin{subfigure}{.49\linewidth}
\captionsetup{font=footnotesize}
\includegraphics[width=\linewidth]{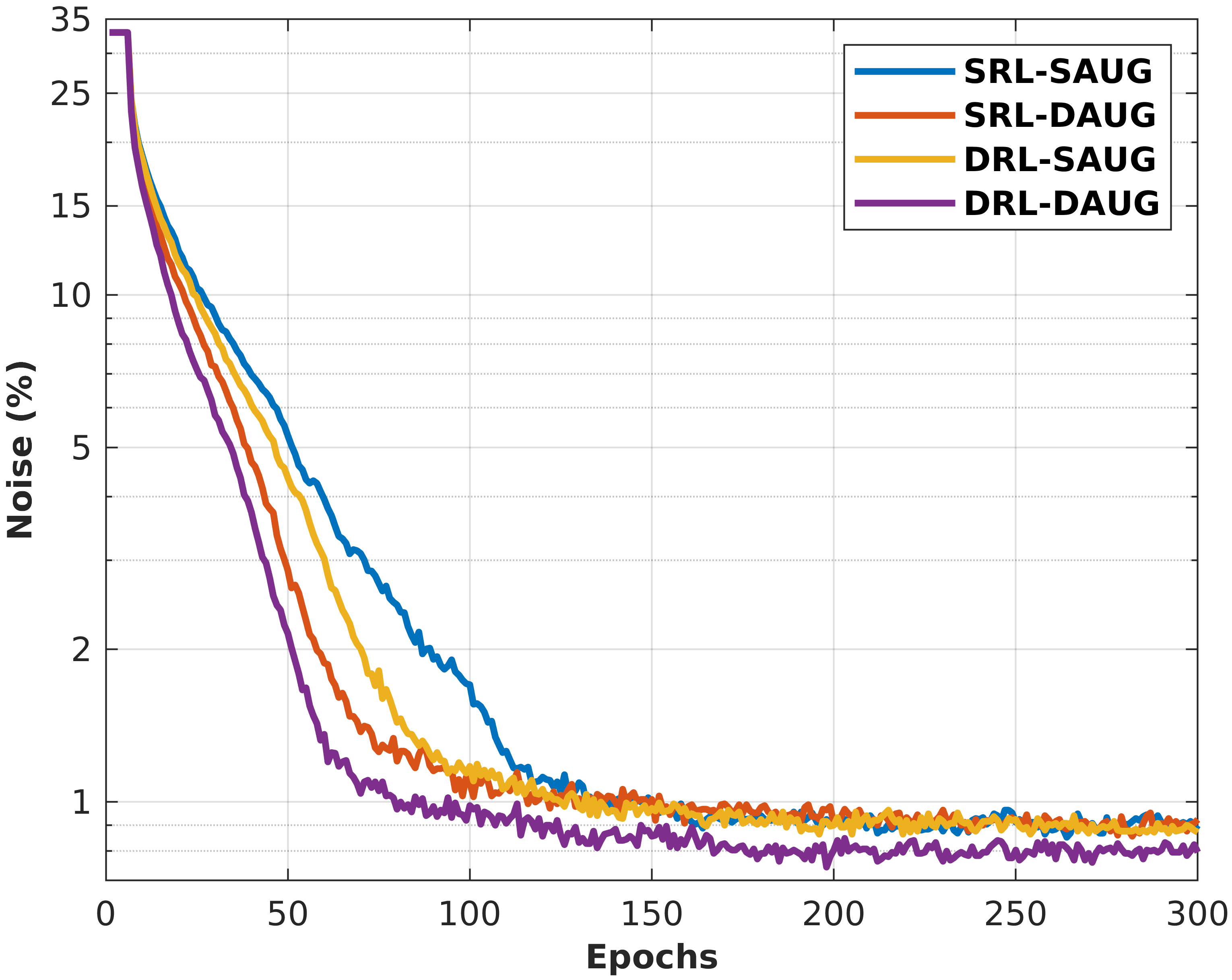}
\caption{}
\label{ablate:sub-4}
\end{subfigure}
\vspace{-0.5em}
\caption{Ablation study considering SVHN$\rightarrow$MNIST UDA task to determine optimal parameters of our proposed AdaPLR method.
(a): Single model is trained with $1$ residual-label (RL) to choose the best $\alpha$ required to compute adaptive noise filtering threshold $\gamma$. (b): Searching for the right number of RL \textit{i.e.,} $N_{RL}$.
(c): Searching for the optimal number of members in the ensemble network ($N_{RL}=4$ is used for $N_e=1$). (d): Investigating the effect of same/disjoint RL (SRL and/or DRL) and same/different data augmentation (SAUG and/or DAUG) in all four possible scenarios. 
}
\vspace{-0.9em}
\label{fig:ablate_mclANDn}
\end{figure}
We ablate the design choices described in section \ref{Method} on the SVHN$\rightarrow$MNIST adaptation task. Parameters found here are then used in all subsequent experiments.

The adaptive property of AdaPLR is a function of $\alpha$. We consider $N_e=1$ (\ie, no ensemble) to ablate the different values of $\alpha$. Figure~\ref{ablate:sub-1} shows that $\alpha=0.90$ results in the highest noise reduction, whereas $\alpha=0.50$ is the least effective. These findings make sense, since with $\alpha=0.50$, the reassignment would start too early (\ie, as soon as some example has confidence greater than 0.5, since $\gamma$ is always zero beforehand), when the network is not yet "ready" for it. Instead, with $\alpha=0.90$, reassignment will start only when some samples have very high confidence. From that moment on, $\gamma$ starts adapting, so that the more the samples with high confidence, the more permissive the threshold $\gamma$ will be. The reason $\alpha=0.95$ results slightly less effective is because a relatively higher number of noisy samples overfits before they are reassigned.

Keeping $N_e=1$ and $\alpha=0.9$, we ablate on the different values of parameter $N_{RL}$ (Figure~\ref{ablate:sub-2}). Results show that $N_{RL}=3,4,$ or $5$ can be considered as the legitimate choice. Further, the results shown in Figure~\ref{ablate:sub-3} exhibit the improvement achieved by the RL approach in comparison to $N_e=1$ (with $N_{RL}=4$, i.e. the best choice found in Figure~\ref{ablate:sub-2}). Though faster noise reduction is achieved with higher number of ensemble members, $N_e=3$ can be regarded as the optimal choice considering performance \textit{vs.} computational cost trade-off. Such improvement is a consequence of the strong consensus obtained from the ensemble network since $N_e-1$ members always receive noiseless feedback, while only one member may be partially affected.

Also, as shown in Figure~\ref{ablate:sub-4}, just one type of augmentation for all the ensemble members (ref. SAUG in the figure) is not enough. 
Also, the results validate the ineffectiveness of using same residual labels (SRL) for all. The best noise reduction is achieved using Disjoint Residual Labels along with different stochastic augmentations, DRL and DAUG in the figure, respectively (for more details see the Appendix.).

\subsection{Results}\label{Results}
The reported results in Table~\ref{tab:digit5}-\ref{tab:domainnet} represent the mean of 3 runs\footnote{Additional details such as standard deviation and remaining noise in refined pseudo-labels are provided in the Appendix}. 
%
In Table~\ref{tab:digit5} \textit{(left)}, we compare AdaPLR with the existing methods which address the challenging \textit{MNIST}$\rightarrow$\textit{SVHN} and \textit{MNIST}$\rightarrow$\textit{MNIST-M} tasks in a multi-target UDA framework.
In both cases, the source contains gray-scale images, and the target holds colored (RGB) images, carrying a massive distribution gap across domains 
for which our method achieves third and second-best performance, respectively. Nevertheless, by outperforming in 3 out of 5 cases, our method achieves state-of-the-art average accuracy. 
For multi-source UDA task in Table~\ref{tab:digit5} \textit{(right)}, the performance of proposed AdaPLR method is slightly affected 
while adapting to \textit{Synthetic-Digits} benchmark. However, in the remaining 4 out of 5 cases, our method outperforms existing methods and achieves state-of-the-art average accuracy. Especially, the difference is substantial in the case of \textit{MNIST-M}.

In Table~\ref{tab:PACS} \textit{(left)}, we compare AdaPLR with the existing methods addressing  multi-target UDA on PACS. As can be noticed, despite the sub-optimal performance in 2 cases, our method achieves superior average accuracy. For multi-source UDA, we compare recent works in Table~\ref{tab:PACS} \textit{(right)}. Also in this framework, our method consistently outperforms existing methods, with only in one case getting lower, but comparable, accuracy. To the best of our knowledge, we are the first to report single-source UDA results on PACS. So, in Table~\ref{tab:PACS_ssda}, we consider similar pairs as of Table~\ref{tab:PACS} \textit{(left)} to evaluate the performance difference. As expected, single-source UDA brings comparatively better performance because of the pair-wise UDA.

In Table~\ref{tab:VisdaC}, along with 2 comparable results for Visda-C, the proposed AdaPLR method achieves superior performance in 6 out of 12 categories that give rise to state-of-the-art average accuracy on such a challenging benchmark. Also In Table~\ref{tab:domainnet}, except one case, AdaPLR consistently outperforms existing methods despite the large number of classes and discrepancy across domains.
  
\paragraph{Discussion.} 
In the context of pseudo-label refinery framework, the single-source and multi-target UDA cases
can be considered as the most challenging tasks since the pre-trained source model is optimized for one particular data distribution only. 
Consequently, inferred pseudo-labels are affected by a comparatively higher amount of shift-noise with respect to the multi-source UDA scenario. Thus, in this case, AdaPLR employs a larger amount of epochs for reducing noise and refining pseudo-labels.
On the other hand, starting from a better pre-trained source model in multi-source UDA, AdaPLR performs better and faster.
Moreover, there is no existing method in the literature that targets all three frameworks (\ie, single-source, multi-target, and multi-source UDA) at a time.
To sum up, AdaPLR outperforms existing methods in all scenarios, even without using source data, which highlights the general applicability of the proposed method to cope with challenging tasks of diverse complexity.

\section{Conclusions}\label{Conclusion}
In this paper, we proposed a novel approach to address source-free UDA, which is cast from the perspective of pseudo-label refinement. To this end, we proposed a novel Ensemble Learning method based on distinct input and feedback by means of Disjoint Residual Labels under the Negative Ensemble Learning framework. Thanks to this new training procedure that allows an extraordinary cleaning of the target data labels, which are sufficient enough to train a single target model by a standard supervised learning method. The proposed approach is found to be quite effective in solving challenging UDA tasks without using source data; it does not require any tuning of parameters and can work in single-source, multi-target, and multi-source scenarios indifferently. Results demonstrate the actual goodness of the proposed approach, outperforming the state-of-the-art average performance in all tested benchmarks.



{\small
\bibliographystyle{ieee_fullname}
\bibliography{egbib}
}

\clearpage

\section*{Appendix}

As mentioned in the main paper, the following sections include:
    (1) Details on the distribution of shift-noise and its impact on negative learning.
    (2) Noise cleaning performance of the proposed method, together with standard deviations for the final accuracy (not included in the main paper due to lack of space).

\subsection*{On the impact of noise distribution on Negative Learning}
\begin{figure*}[!t]
\captionsetup{font=small}
\centering
\hspace{0.2em}
\begin{subfigure}{0.3\textwidth}
\centering
\captionsetup{font=small}
\includegraphics[width=0.8\linewidth]{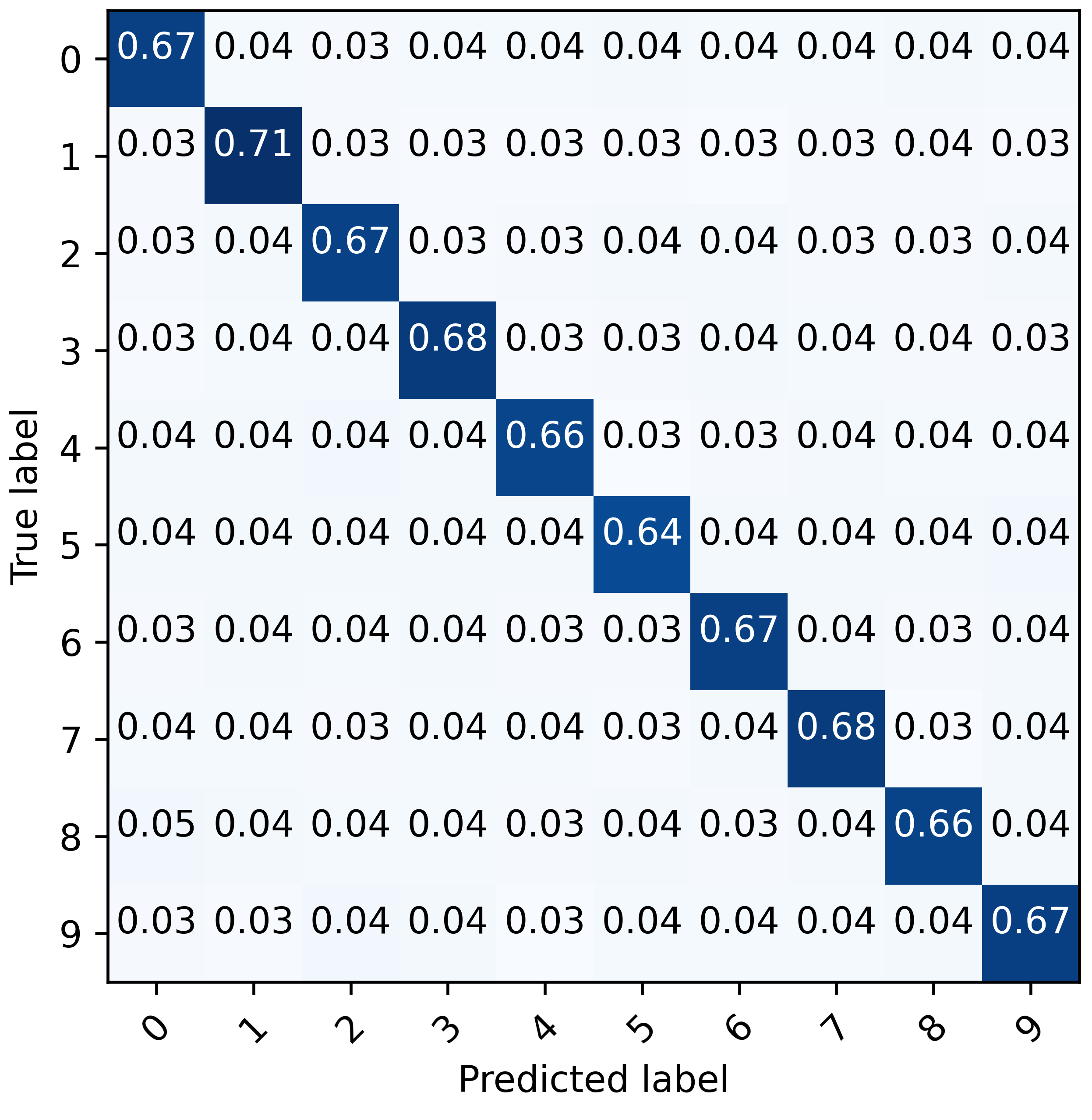}
\end{subfigure}
\hspace{0.6em}
\begin{subfigure}{0.3\textwidth}
\centering
\captionsetup{font=small}
\includegraphics[width=0.8\linewidth]{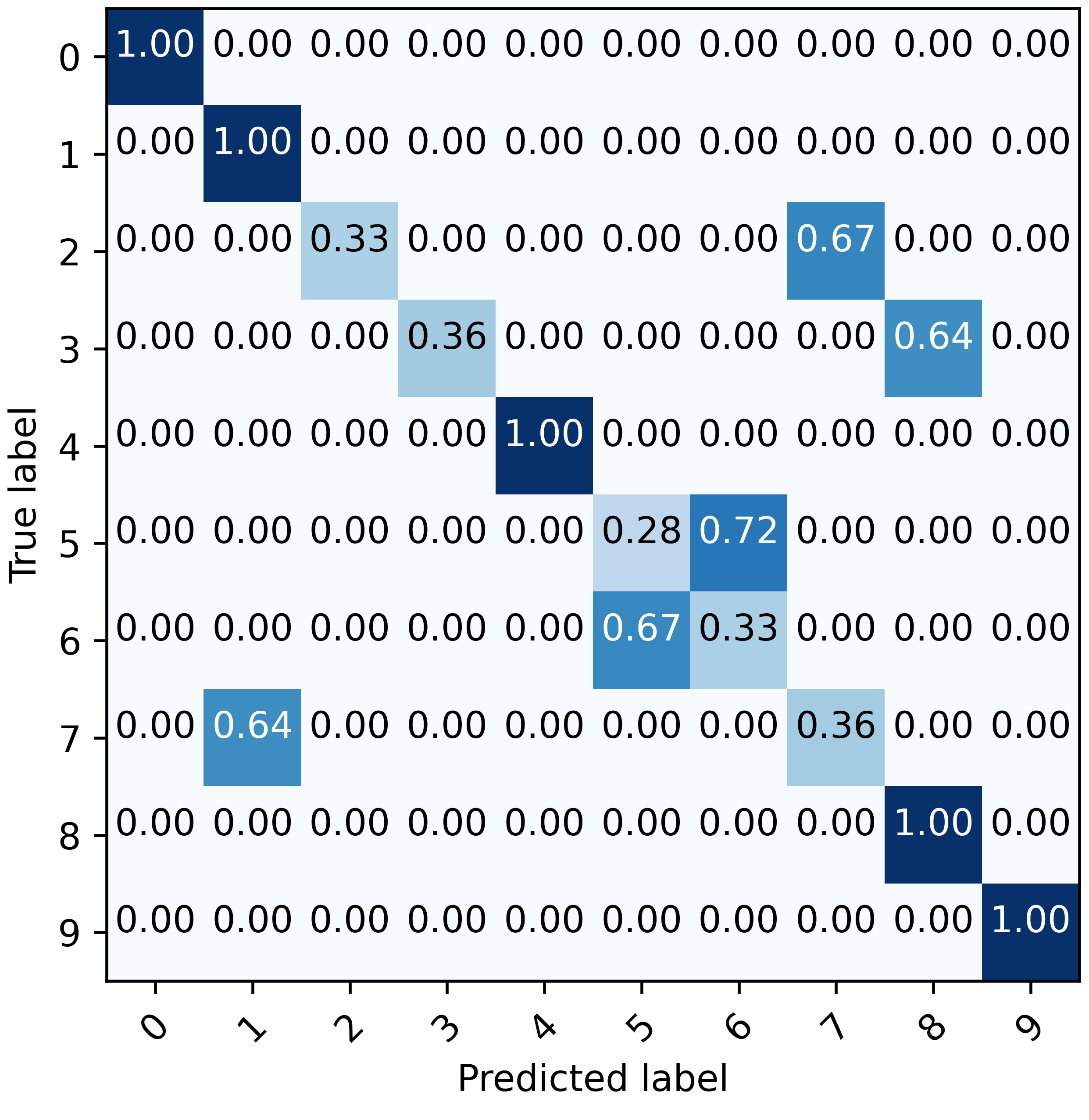}
\end{subfigure}
\hspace{0.6em}
\begin{subfigure}{0.3\textwidth}
\centering
\captionsetup{font=small}
\includegraphics[width=0.8\linewidth]{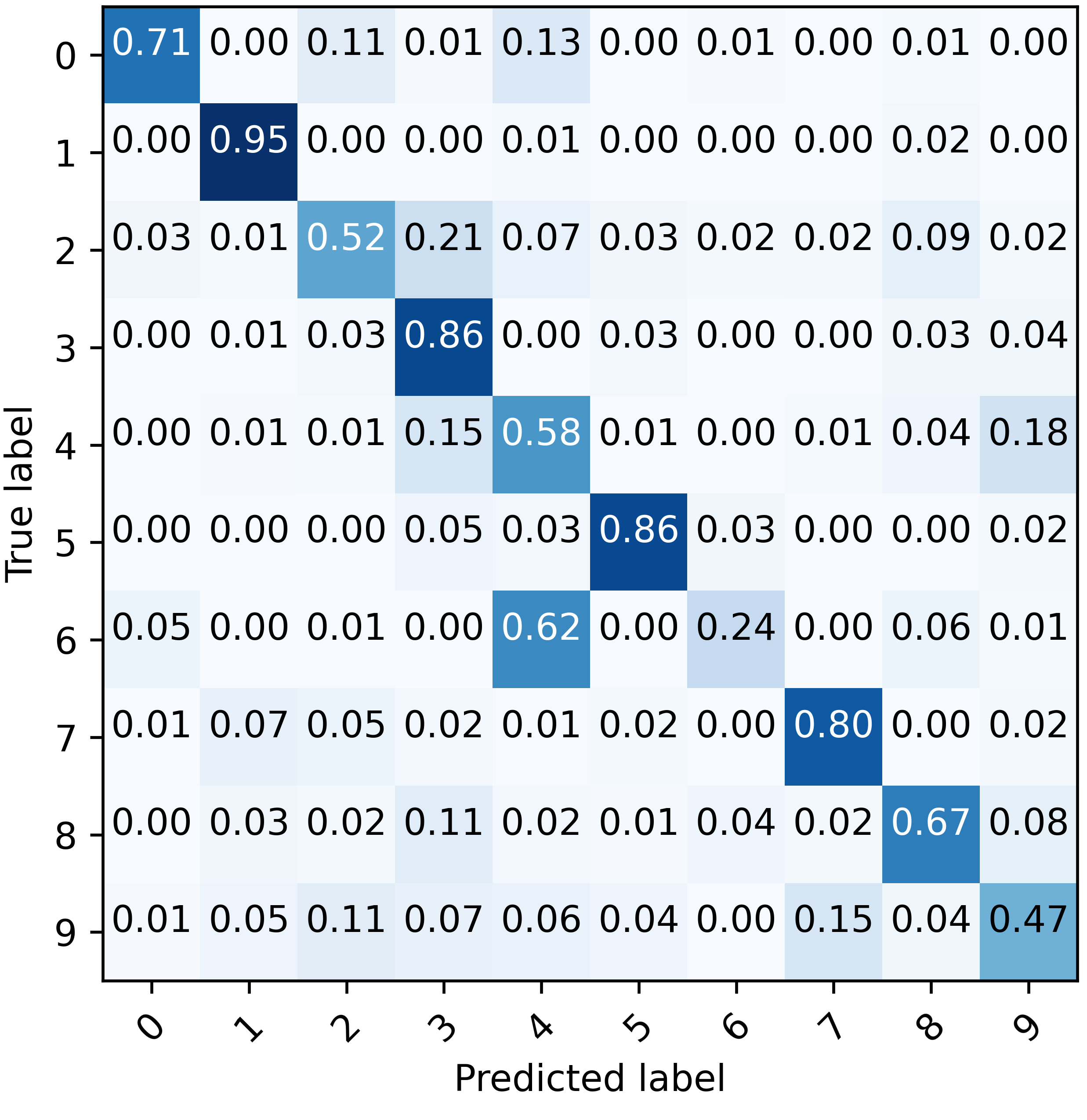}
\end{subfigure}
\\
\begin{subfigure}{0.31\textwidth}
\centering
\captionsetup{font=small}
\includegraphics[width=\linewidth]{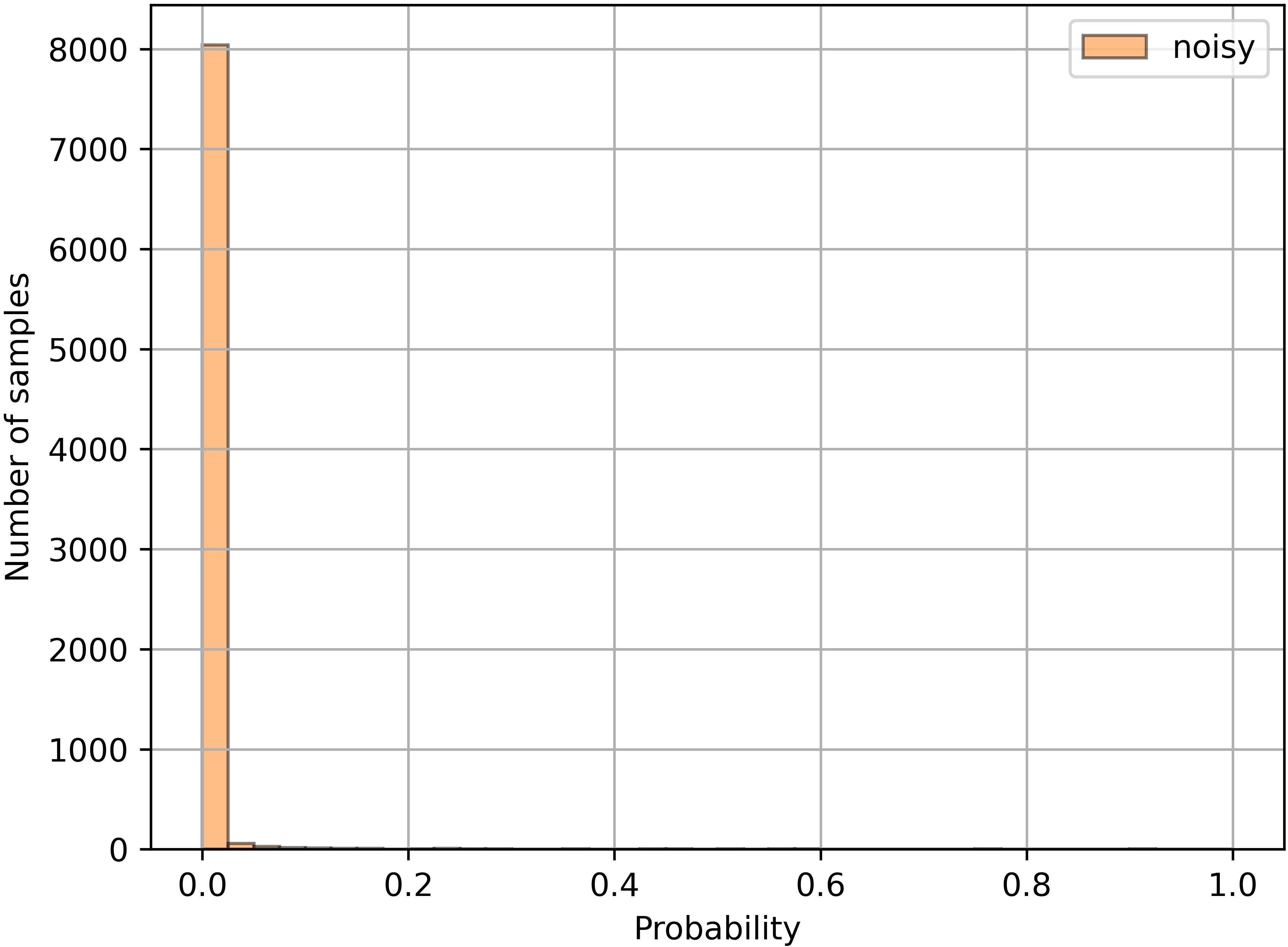}
\end{subfigure}
\hspace{0.3em}
\begin{subfigure}{0.305\textwidth}
\centering
\captionsetup{font=small}
\includegraphics[width=\linewidth]{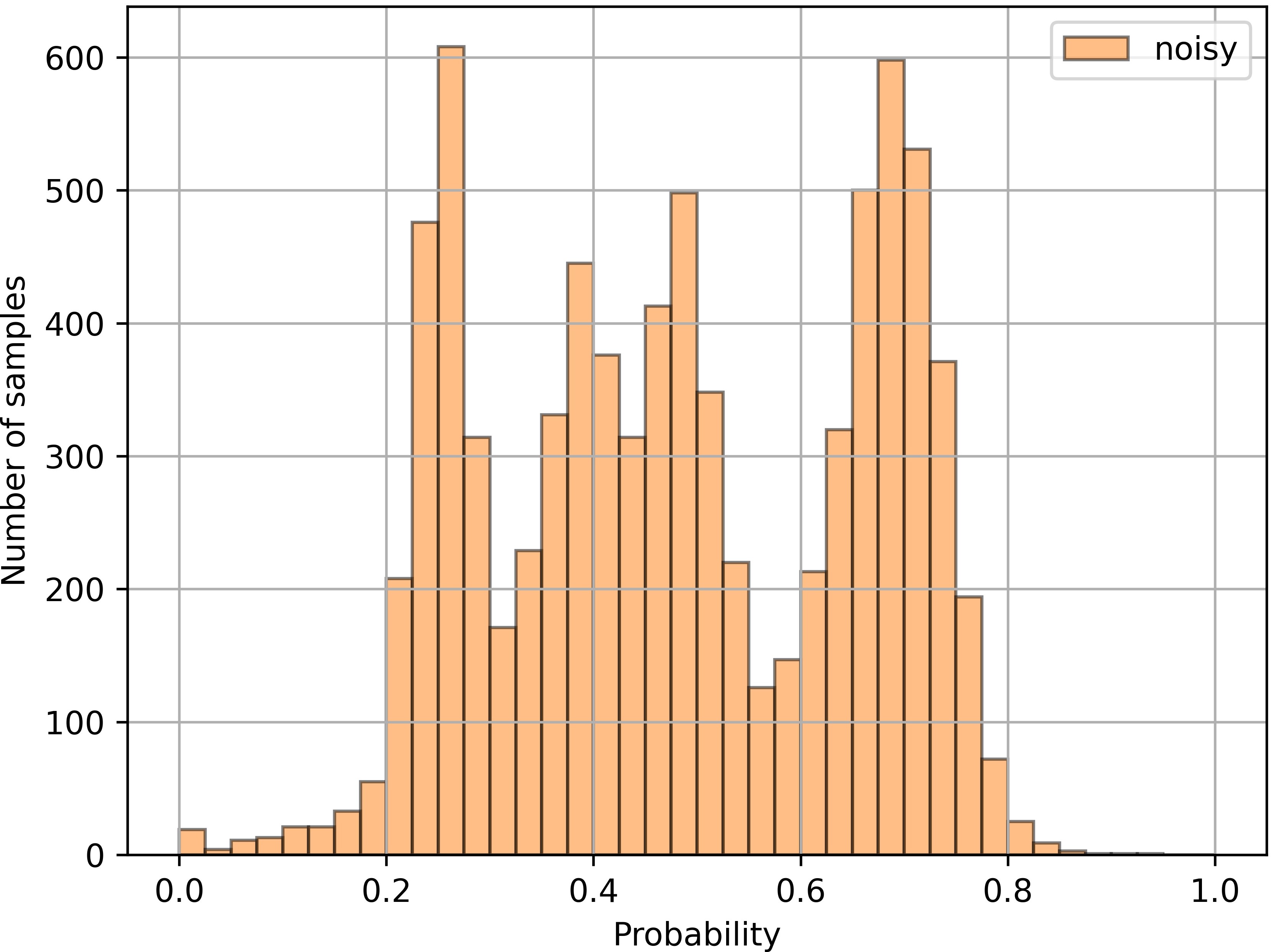}
\end{subfigure}
\hspace{0.3em}
\begin{subfigure}{0.305\textwidth}
\centering
\captionsetup{font=small}
\includegraphics[width=\linewidth]{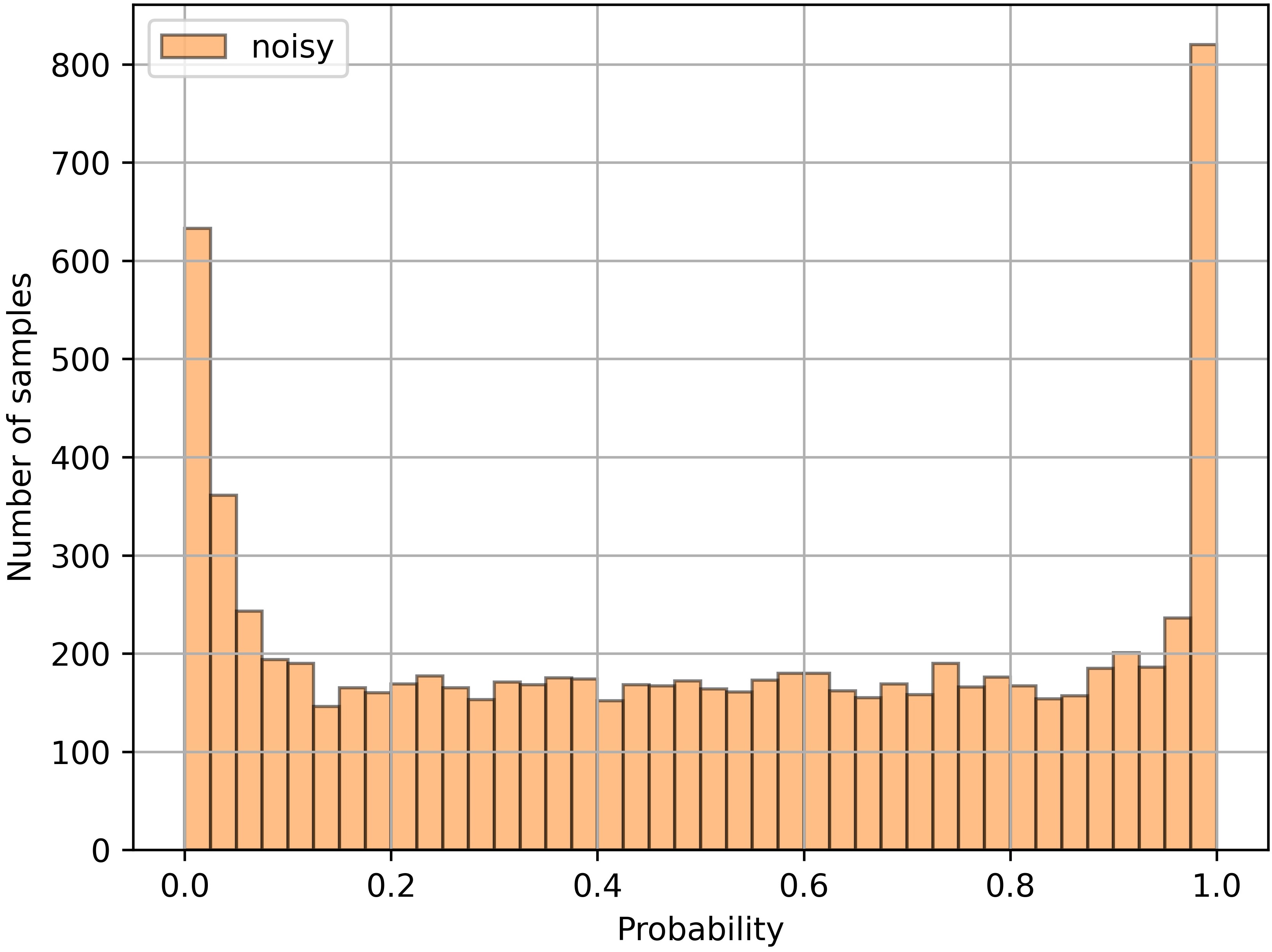}
\end{subfigure}
\\
\begin{subfigure}{0.315\textwidth}
\centering
\captionsetup{font=small}
\includegraphics[width=\linewidth]{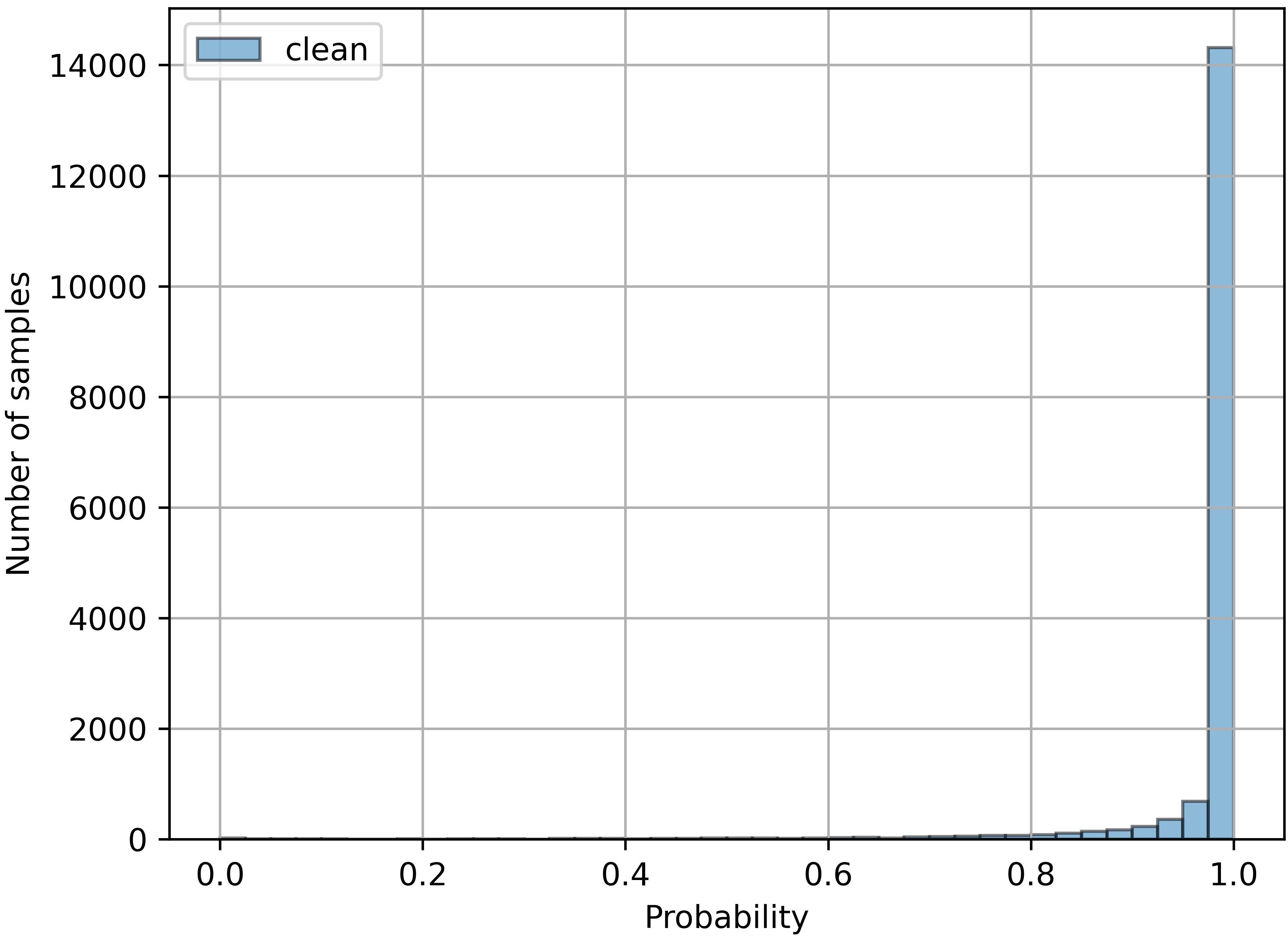}
\caption{Symmetric-Noise}
\label{NL:symm}
\end{subfigure}
\begin{subfigure}{0.317\textwidth}
\centering
\captionsetup{font=small}
\includegraphics[width=\linewidth]{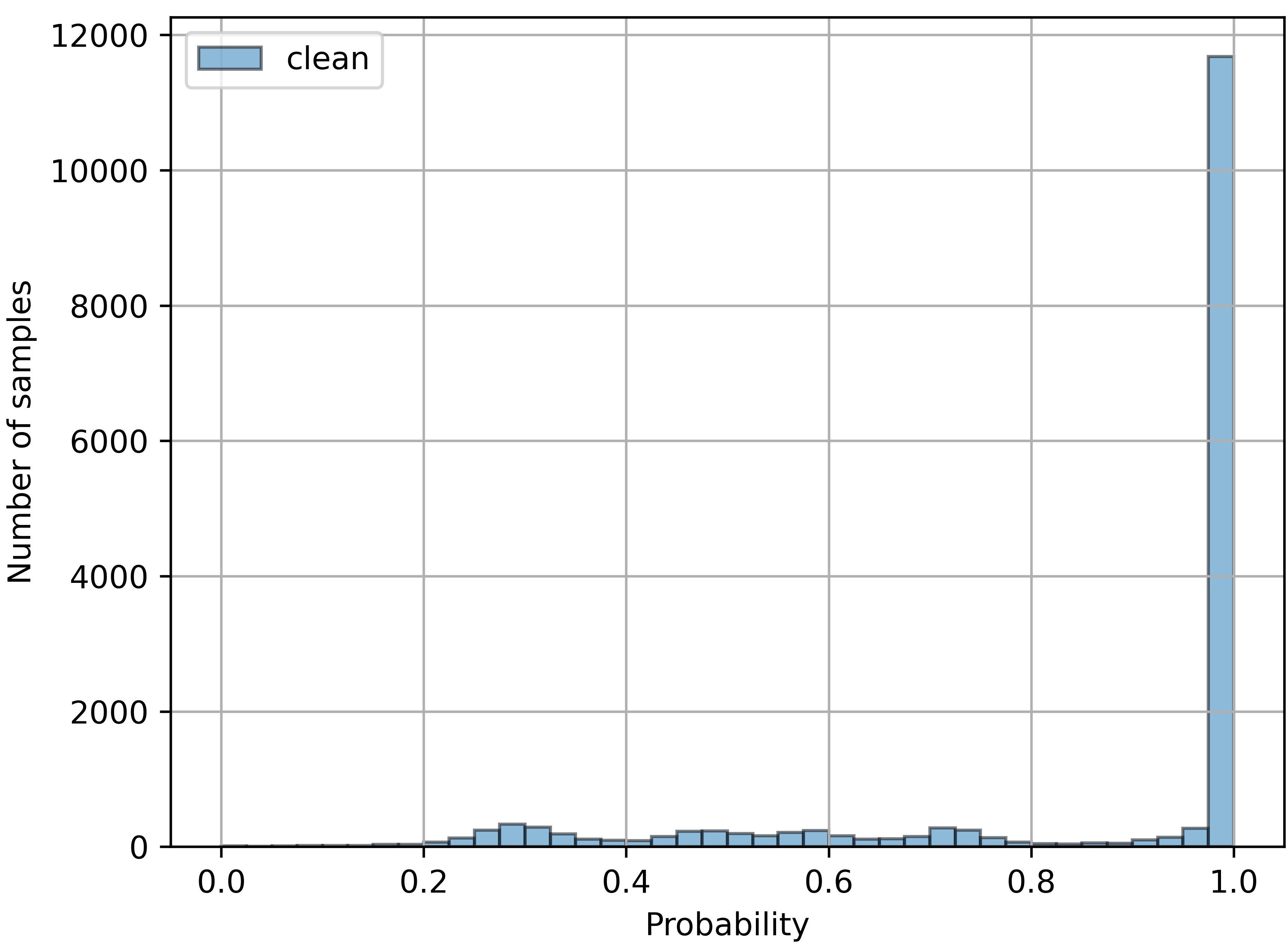}
\caption{Asymmetric-Noise}
\label{NL:asymm}
\end{subfigure}
\begin{subfigure}{0.317\textwidth}
\centering
\captionsetup{font=small}
\includegraphics[width=\linewidth]{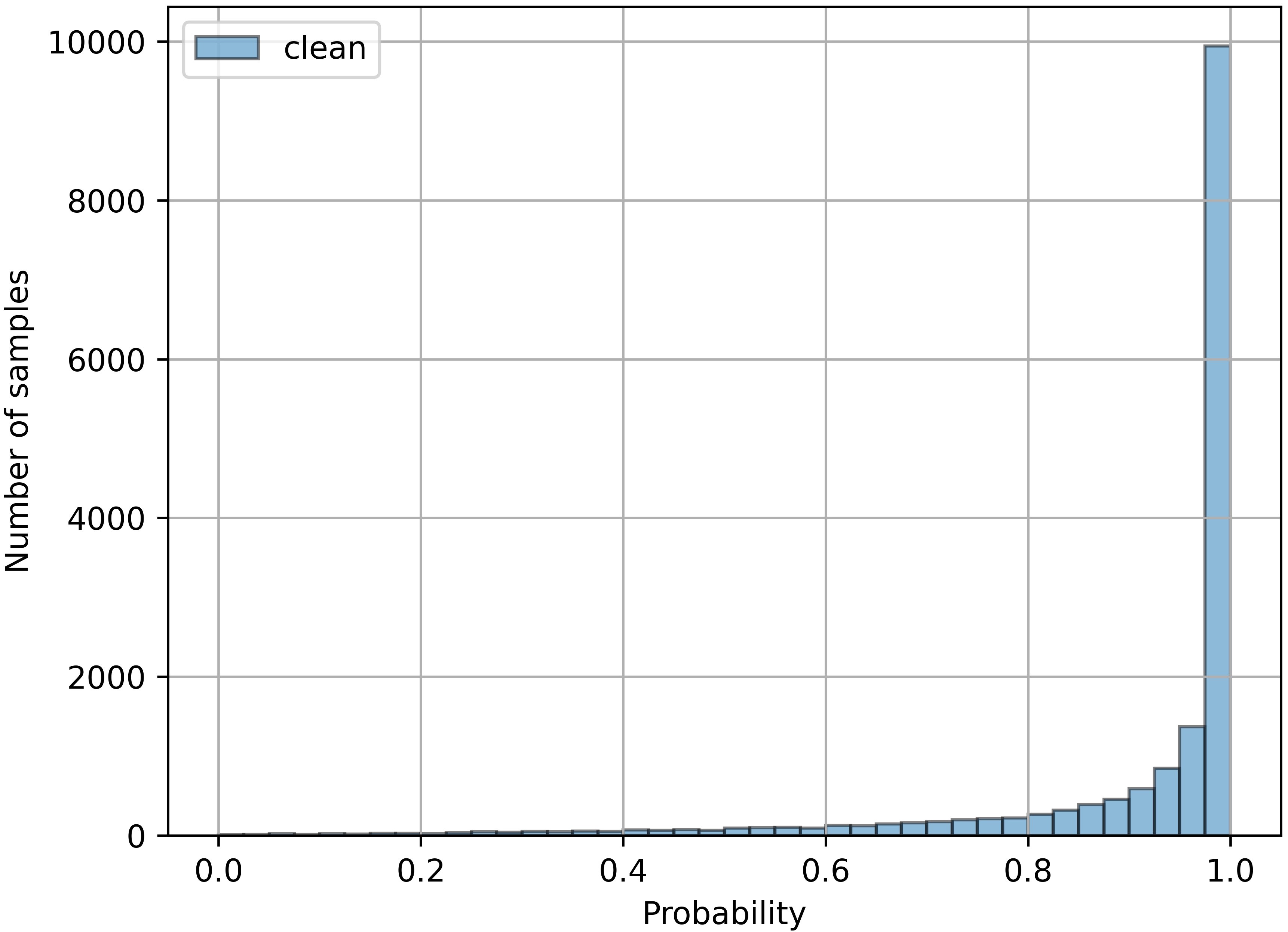}
\caption{Shift-Noise}
\label{NL:shift}
\end{subfigure}
%
\caption{Noise filtering capability of the existing Negative Learning method \cite{kim2019nlnl} over various noise distributions. Column (a) Symmetric Noise, column (b) Asymmetric artificial noise \cite{patrini2017making}, column (c) Shift noise \cite{morerio2020generative}. First row:  the confusion matrix shows how the noise is distributed in the beginning. Second row: confidence prediction for the noisy samples after training with NL. Third row: confidence prediction for the clean samples after training with NL. The amount of initial noise is same in magnitude (\ie 32.97\%) for all the cases.}
\label{NL:all}
\end{figure*}

Negative Learning (NL) is an indirect learning method in which a model is optimized to produce the lowest confidence to a randomly chosen complementary label for the given input image. Such a method produces more reliable performance in the case of a noisy label set.
Nevertheless, the applicability of the existing NL method \cite{kim2019nlnl} is limited to a type of noise showing uniform distribution (\textit{Symmetric-Noise}). In order to highlight specific limitations of the existing NL method, figure \ref{NL:all}  shows the confusion matrix for the initial noise along with the histograms of the noisy and clean sample's confidence distribution obtained after training for SVHN $\to$ MNIST UDA. 

As shown in Figure~\ref{NL:symm}, when labels are initially affected by \textit{Symmetric-Noise}, noisy samples are classified with low confidence whereas the clean samples are leaned to high confidence. Consequently, effective noise separation is achieved.

For the \textit{Asymmetric-Noise}, mimicking some of the structures of real errors \cite{patrini2017making} \textit{i.e.,} for MNIST, mapping 2$\to$7, 3$\to$8, 7$\to$1, and 5$\leftrightarrow$6, existing NL method's effectiveness degrades considerably. As can be seen in Figure~\ref{NL:asymm}, noisy samples are overfitted with quite high confidence (even higher than 50\%).
Yet, sub-optimal separation still exists if we consider only samples carrying confidence higher than 90\% (though it is not generalizable for other benchmarks where all clean samples do not carry such high confidence). 

Nevertheless, noisy samples are overfitted with very high confidence when \textit{shift-noise}  associated with inferred pseudo-labels \cite{morerio2020generative} is considered (Figure~\ref{NL:shift}). Consequently, subsequent Positive Learning achieves sub-optimal performance with such noise distribution in the UDA framework.

Please note that in all cases, the \textit{amount} of noise is same \ie 32.97\%, similar to the amount of shift-noise observed in the case of SVHN $\to$ MNIST UDA task in Table \ref{supp:digit5}. 

\subsection*{Noise Cleaning Performance of AdaPLR}
\begin{figure*}[ht]
\captionsetup{font=small}
\centering
\begin{subfigure}{0.3\textwidth}
\centering
\captionsetup{font=small}
\includegraphics[width=\textwidth]{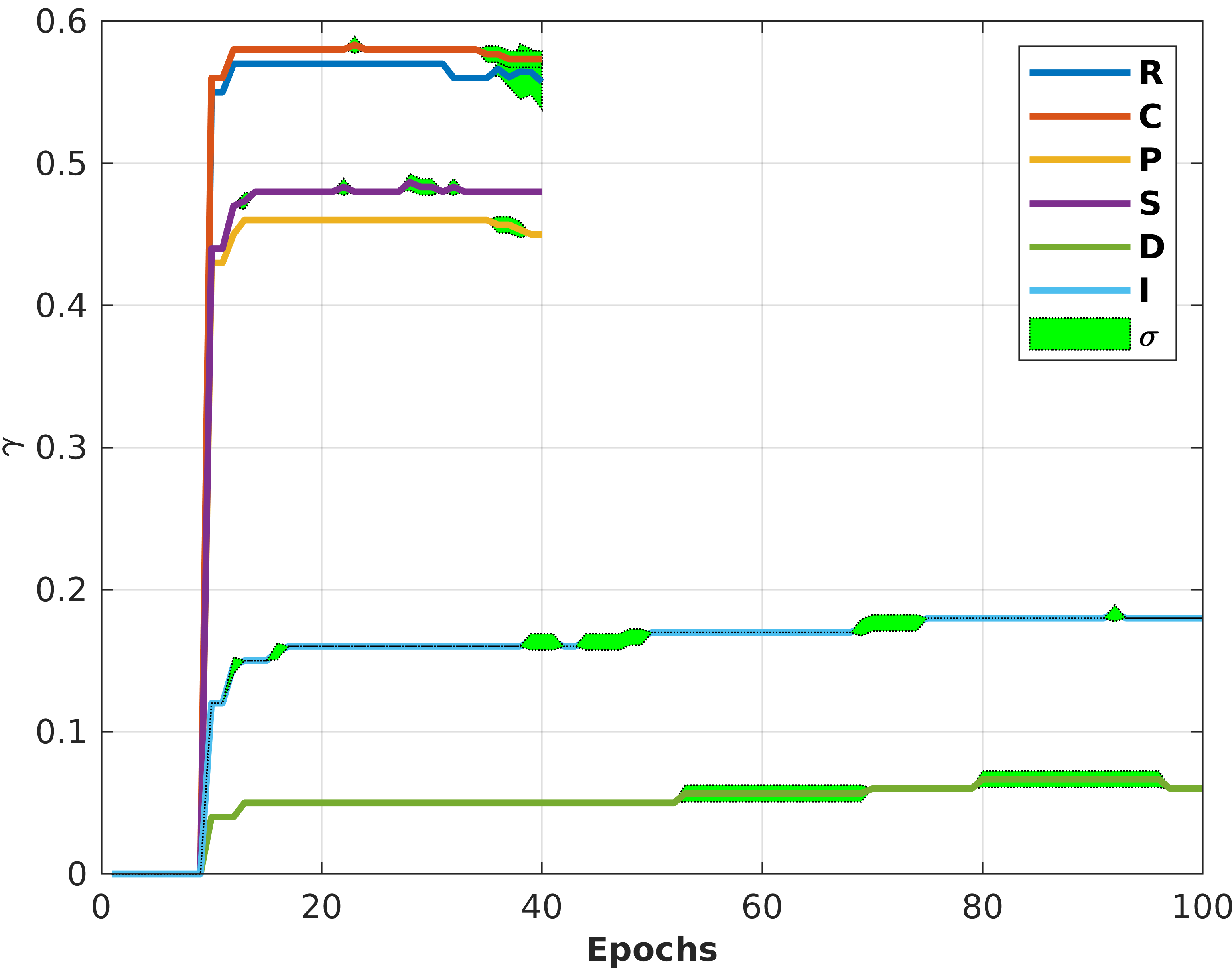}
\caption{Adaptiveness of $\gamma$ threshold.}
\label{DN:gamma}
\end{subfigure}
\begin{subfigure}{0.69\textwidth}
\centering
\captionsetup{font=small}
\includegraphics[width=\textwidth]{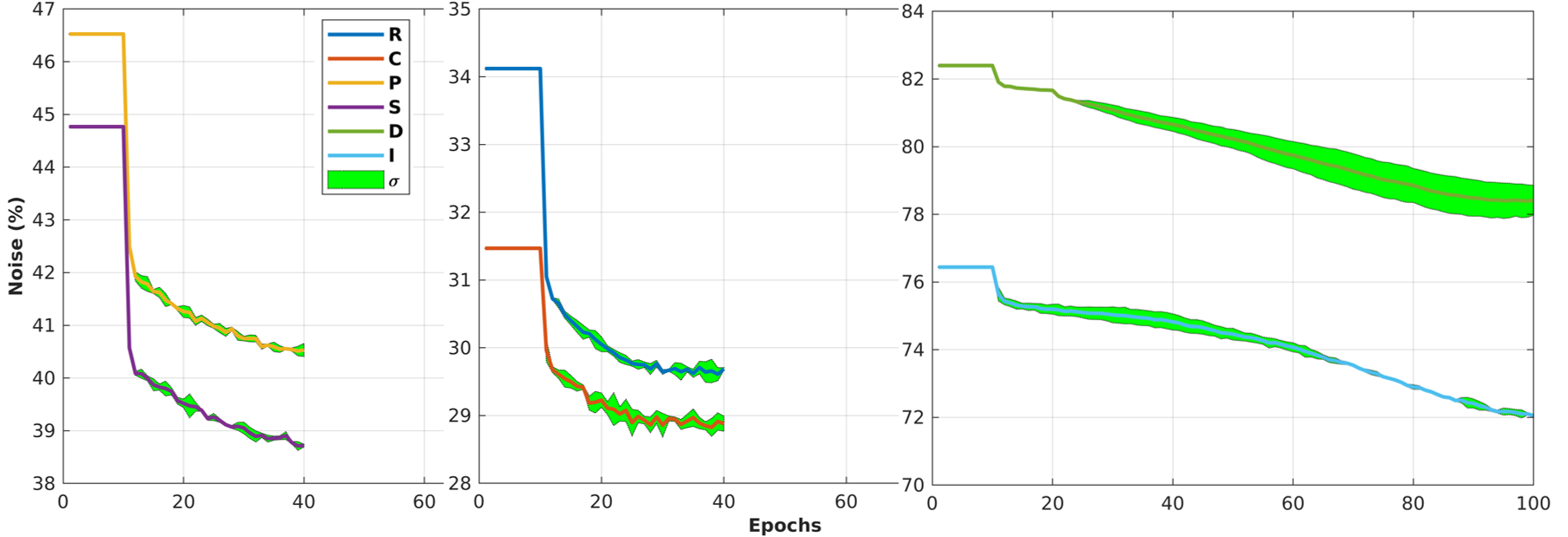}
\caption{Progressive noise reduction by pseudo-label refinement.}
\label{DN:noise}
\end{subfigure}
%
\caption{Training evolution on DomainNet. Multi-source case: for each target, the rest of the domains are considered as source. For better representation, we concatenate noise reduction trends with different scales in (b). 
Legend: \textit{\textcolor{blue}{\textbf{C}}: Clipart, \textcolor{blue}{\textbf{I}}: Infograph, \textcolor{blue}{\textbf{P}}: Painting, \textcolor{blue}{\textbf{Q}}: Quickdraw, \textcolor{blue}{\textbf{R}}: Real, \textcolor{blue}{\textbf{S}}: Sketch, and \textcolor{blue}{\textbf{$\sigma$}}: Instantaneous standard deviation of three runs.}
}
\label{DN:all}
\end{figure*}

We demonstrate here the adaptive noise filtering and progressive pseudo-label refinement performance of the proposed AdaPLR method. In Figure~\ref{DN:all} we evaluate the robustness of our method in achieving effective pseudo-label refinement over three runs while considering the most challenging UDA task \ie, multi-source UDA on DomainNet. 
Specifically, Figure~\ref{DN:gamma} shows the adaptiveness of $\gamma$ threshold for different UDA tasks in which inferred pseudo-labels are affected by the various amount of noise \ie, starting from 31.47\% up to 82.40\% of shift-noise. As can be seen in Figure~\ref{DN:noise}, in each case, AdaPLR achieves quite reasonable noise reduction throughout the training process. 
Such a trend is also observed in single-source UDA on \textit{Digit5} benchmarks (see main paper). However, the only difference in the two cases is that the change in $\gamma$ threshold is comparatively smoother for Digit5 which eventually takes more epochs for pseudo-label refinement (and achieves better noise reduction too).

To conclude, in Table~\ref{supp:digit5}-\ref{supp:domainnet}, we summarize statistics concerning classification accuracy along with standard deviations of \textit{inferred} pseudo-labels --- obtained using pre-trained source model, \textit{refined} pseudo-labels --- obtained using the proposed Negative Ensemble Learning (NEL) technique, and the single-target model trained with the refined pseudo-labels --- the final stage of the proposed \textit{AdaPLR} method. In many cases, we achieve better performance with NEL only. However, for a fair comparison with existing works, the main paper compares the results achieved using a single target model only. Yet, the reported results validate that the proposed AdaPLR is remarkably effective and stable for all the UDA benchmarks.


\begin{table*}[!h]
\centering
\begin{subtable}{\textwidth}
\centering
\small
\begin{tabular}{|l|c c c c c |c|}
\hline
Source & 
\textcolor{blue}{\textbf{\textit{T}}} &
\textcolor{blue}{\textbf{\textit{T}}} &
\textcolor{blue}{\textbf{\textit{T}}} &
\textcolor{blue}{\textbf{\textit{S}}} &
\textcolor{blue}{\textbf{\textit{U}}} &
\multirow{2}{*}{Avg.} \\
\cline{1-6}
Target & 
\textcolor{blue}{\textbf{\textit{U}}} &
\textcolor{blue}{\textbf{\textit{S}}} &
\textcolor{blue}{\textbf{\textit{M}}} &
\textcolor{blue}{\textbf{\textit{T}}} &
\textcolor{blue}{\textbf{\textit{T}}} &
 \\
\hline
Inferred   & 86.3 & 34.7 & 63.8 & 67.0 & 70.2 & 64.4 \\
Refined    & 99.1$\pm$0.04 & 62.0$\pm$0.10 & 97.5$\pm$0.03 & 99.2$\pm$0.05 & 99.2$\pm$0.02 & 91.4$\pm$0.05 \\
AdaPLR & 97.4$\pm$0.10 & 61.6$\pm$0.29 & 95.4$\pm$0.16 & 99.2$\pm$0.02 & 99.2$\pm$0.02 & 90.6$\pm$0.12 \\
\hline
\end{tabular}
\caption{Single-Source UDA.}
\label{supp:ssda-digit5}
\end{subtable}
\begin{subtable}{\textwidth}
\centering
\small
\begin{tabular}{|l|c c c c c |c|}
\hline
Source & 
\textcolor{blue}{\textbf{\textit{M,S,D,U}}} &
\textcolor{blue}{\textbf{\textit{T,S,D,U}}} &
\textcolor{blue}{\textbf{\textit{T,M,D,U}}} &
\textcolor{blue}{\textbf{\textit{T,M,S,U}}} &
\textcolor{blue}{\textbf{\textit{T,M,S,D}}} &
\multirow{2}{*}{Avg.} \\
\cline{1-6}
Target & 
\textcolor{blue}{\textbf{\textit{T}}} &
\textcolor{blue}{\textbf{\textit{M}}} &
\textcolor{blue}{\textbf{\textit{S}}} &
\textcolor{blue}{\textbf{\textit{D}}} &
\textcolor{blue}{\textbf{\textit{U}}} &
 \\
\hline
Inferred   & 98.6 & 69.1 & 52.0 & 40.3 & 88.7 & 69.8 \\
Refined   & 98.8$\pm$0.95 & 94.2$\pm$0.18 & 84.6$\pm$0.66 & 87.8$\pm$0.85 & 98.6$\pm$0.03 & 92.8$\pm$0.53 \\
AdaPLR & 99.1$\pm$0.02 & 95.5$\pm$0.71 & 89.6$\pm$0.55 & 90.0$\pm$0.63 & 97.8$\pm$0.15 & 94.4$\pm$0.41 \\
\hline
\end{tabular}
\caption{Multi-Source UDA.}
\label{supp:msda-digit5}
\end{subtable}
\caption{Results on Digit5. 
Legend: \textit{\textcolor{blue}{\textbf{T}}: MNIST, \textcolor{blue}{\textbf{S}}: SVHN, \textcolor{blue}{\textbf{U}}: USPS, \textcolor{blue}{\textbf{M}}: MNIST-M, and \textcolor{blue}{\textbf{D}}: Synthetic-Digits.}
}
\label{supp:digit5}
\end{table*}

\begin{table*}[!h]
\centering
\small
\begin{subtable}{\textwidth}
\centering
\small
\begin{tabular}{|l|c c c | c c c |c|}
\hline
Source & 
\textcolor{blue}{\textbf{\textit{ }}} &
\textcolor{blue}{\textbf{\textit{P}}} &
\textcolor{blue}{\textbf{\textit{ }}} &
\textcolor{blue}{\textbf{\textit{ }}} &
\textcolor{blue}{\textbf{\textit{A}}} &
\textcolor{blue}{\textbf{\textit{ }}} &
Avg. \\
\hline
\hline
Target (Combined) & 
\textcolor{blue}{\textbf{\textit{ }}} &
\textcolor{blue}{\textbf{\textit{A,C,S}}} &
\textcolor{blue}{\textbf{\textit{ }}} &
\textcolor{blue}{\textbf{\textit{ }}} &
\textcolor{blue}{\textbf{\textit{P,C,S}}} &
\textcolor{blue}{\textbf{\textit{ }}} &
 \\
\hline
Inferred   & & 37.7 & & & 57.9 & & 47.8 \\
Refined   & & 57.3$\pm$1.13 & & & 73.8$\pm$0.81 & & 65.6$\pm$0.97 \\
\hline
\hline
Target & 
\textcolor{blue}{\textbf{\textit{A}}} &
\textcolor{blue}{\textbf{\textit{C}}} &
\textcolor{blue}{\textbf{\textit{S}}} &
\textcolor{blue}{\textbf{\textit{P}}} &
\textcolor{blue}{\textbf{\textit{C}}} &
\textcolor{blue}{\textbf{\textit{S}}} &
 \\
\hline
AdaPLR & 80.1$\pm$0.37 & 76.1$\pm$1.62 & 25.9$\pm$0.82 & 96.0$\pm$0.34 & 82.8$\pm$1.09 & 49.8$\pm$0.89 & 68.4$\pm$0.90 \\
\hline
\end{tabular}
\caption{Multi-Target UDA. The final accuracy on each target is achieved using the same target model trained with refined pseudo-labels.}
\end{subtable}
\begin{subtable}{\textwidth}
\centering
\small
\begin{tabular}{|l|c c c c c c |c|}
\hline
Source & 
\textcolor{blue}{\textbf{\textit{P}}} &
\textcolor{blue}{\textbf{\textit{P}}} &
\textcolor{blue}{\textbf{\textit{P}}} &
\textcolor{blue}{\textbf{\textit{A}}} &
\textcolor{blue}{\textbf{\textit{A}}} &
\textcolor{blue}{\textbf{\textit{A}}} &
\multirow{2}{*}{Avg.} \\
\cline{1-7}
Target & 
\textcolor{blue}{\textbf{\textit{A}}} &
\textcolor{blue}{\textbf{\textit{C}}} &
\textcolor{blue}{\textbf{\textit{S}}} &
\textcolor{blue}{\textbf{\textit{P}}} &
\textcolor{blue}{\textbf{\textit{C}}} &
\textcolor{blue}{\textbf{\textit{S}}} &
 \\
\hline
Inferred   & 60.9 & 24.8 & 26.5 & 96.0 & 58.1 & 43.9 & 51.7 \\
Refined   & 81.1$\pm$0.44 & 76.9$\pm$1.36 & 31.7$\pm$0.98 & 98.2$\pm$0.19 & 80.9$\pm$0.77 & 51.2$\pm$1.04 & 70.0$\pm$0.80 \\
AdaPLR & 82.6$\pm$0.83 & 80.5$\pm$2.66 & 32.3$\pm$0.68 & 98.4$\pm$0.03 & 84.3$\pm$1.50 & 56.1$\pm$1.27 & 72.4$\pm$1.16 \\
\hline
\end{tabular}
\caption{Single-Source UDA.}
\label{supp:ssda-PACS}
\end{subtable}
\begin{subtable}{\textwidth}
\centering
\small
\begin{tabular}{|l|c c c c|c|}
\hline
Source & 
\textcolor{blue}{\textbf{\textit{C,P,S}}} &
\textcolor{blue}{\textbf{\textit{A,P,S}}} &
\textcolor{blue}{\textbf{\textit{A,C,S}}} &
\textcolor{blue}{\textbf{\textit{A,C,P}}} &
\multirow{2}{*}{Avg.} \\
\cline{1-5}
Target & 
\textcolor{blue}{\textbf{\textit{A}}} &
\textcolor{blue}{\textbf{\textit{C}}} &
\textcolor{blue}{\textbf{\textit{P}}} &
\textcolor{blue}{\textbf{\textit{S}}} &
 \\
\hline
Inferred   & 78.4 & 77.9 & 95.3 & 64.5 & 79.0 \\
Refined   & 89.3$\pm$0.36 & 87.2$\pm$0.15 & 98.1$\pm$0.23 & 83.2$\pm$0.86 & 89.5$\pm$0.40 \\
AdaPLR & 90.8$\pm$0.08 & 89.5$\pm$0.54 & 98.8$\pm$0.12 & 85.2$\pm$0.73 & 91.1$\pm$0.37 \\
\hline
\end{tabular}
\caption{Multi-Source UDA.}
\label{supp:msda-PACS}
\end{subtable}
\caption{Results on PACS. 
Legend:  \textit{\textcolor{blue}{\textbf{A}}: Art-Painting, \textcolor{blue}{\textbf{C}}: Cartoon, \textcolor{blue}{\textbf{P}}: Photo, and \textcolor{blue}{\textbf{S}}: Sketch.}
}
\label{supp:PACS}
\end{table*}

\begin{table*}[!h]
\centering
\small
\resizebox{\textwidth}{!}{
\begin{tabular}{|l|c c c c c c c c c c c c |c|}
\hline
Methods & \textcolor{blue}{\textbf{\textit{plane}}} & \textcolor{blue}{\textbf{\textit{bcycl}}} & \textcolor{blue}{\textbf{\textit{bus}}} & \textcolor{blue}{\textbf{\textit{car}}} & \textcolor{blue}{\textbf{\textit{horse}}} & \textcolor{blue}{\textbf{\textit{knife}}} & \textcolor{blue}{\textbf{\textit{mcycl}}} & \textcolor{blue}{\textbf{\textit{person}}} & \textcolor{blue}{\textbf{\textit{plant}}} & \textcolor{blue}{\textbf{\textit{skate}}} & \textcolor{blue}{\textbf{\textit{train}}} & \textcolor{blue}{\textbf{\textit{truck}}} & Avg.  \\
\hline
Inferred & 64.2 & 6.3 & 75.2 & 21.7 & 55.9 & 95.7 & 22.8 & 1.4 & 79.8 & 0.7 & 82.8 & 19.8 & 46.3 \\
Refined & 95.2 & 64.8 & 90.8 & 89.7 & 87.4 & 93.7 & 91.5 & 88.5 & 56.4 & 82.9 & 97.1 & 93.8 & 85.1 \\
  & $\pm$0.05 & $\pm$0.11 & $\pm$0.23 & $\pm$0.27 & $\pm$0.08 & $\pm$0.51 & $\pm$0.21 & $\pm$0.66 & $\pm$1.01 & $\pm$0.37 & $\pm$0.09 & $\pm$0.15 & $\pm$0.31 \\
AdaPLR & 94.5 & 60.8 & 92.3 & 87.3 & 87.3 & 93.2 & 87.6 & 91.1 & 56.9 & 83.4 & 93.7 & 86.6 & 84.2 \\
& $\pm$0.29 & $\pm$0.31 & $\pm$0.46 & $\pm$0.78 & $\pm$0.55 & $\pm$0.02 & $\pm$0.58 & $\pm$0.27 & $\pm$0.09 & $\pm$0.44 & $\pm$0.07 & $\pm$0.74 & $\pm$0.38 \\
\hline
\end{tabular}
}
\caption{Results on Visda-C.}
\label{supp:VisdaC}
\end{table*}

\begin{table*}[!t]
\centering
\captionsetup{width=.85\linewidth}
\small
\begin{tabular}{|l|c c c c c c |c|}
\hline
Target & 
\textcolor{blue}{\textbf{\textit{C}}} &
\textcolor{blue}{\textbf{\textit{I}}} &
\textcolor{blue}{\textbf{\textit{P}}} &
\textcolor{blue}{\textbf{\textit{Q}}} &
\textcolor{blue}{\textbf{\textit{R}}} &
\textcolor{blue}{\textbf{\textit{S}}} &
Avg.
\\
\hline
Inferred & 68.5 & 23.6 & 53.5 & 17.6 & 65.9 & 55.2 & 47.4 \\
Refined & 71.1$\pm$0.11 & 28.0$\pm$0.05 & 59.5$\pm$0.12 & 21.6$\pm$0.44 & 70.4$\pm$0.04 & 61.3$\pm$0.03 & 52.0$\pm$0.13 \\
AdaPLR & 68.3$\pm$0.15 & 22.1$\pm$0.17 & 54.7$\pm$0.15 & 22.8$\pm$0.45 & 67.3$\pm$0.92 & 57.1$\pm$0.27 & 48.7$\pm$0.35 \\
\hline
\end{tabular}
\caption{Multi-Source UDA results on DomainNet. 
Legend: \textit{\textcolor{blue}{\textbf{C}}: Clipart, \textcolor{blue}{\textbf{I}}: Infograph, \textcolor{blue}{\textbf{P}}: Painting, \textcolor{blue}{\textbf{Q}}: Quickdraw, \textcolor{blue}{\textbf{R}}: Real, and \textcolor{blue}{\textbf{S}}: Sketch.}
}
\label{supp:domainnet}
\end{table*}

\end{document}